\documentclass[sigconf]{acmart}
\usepackage{algorithm}
\usepackage{algpseudocode}
\usepackage{ragged2e}
\usepackage{subcaption} %
\usepackage{graphicx}
\usepackage{multirow}
\usepackage{tikz}
\usepackage{xcolor}

\AtBeginDocument{%
  }

\copyrightyear{2025}
\acmYear{2025}
\setcopyright{acmlicensed}\acmConference[ICMR '25]{Proceedings of the 2025
International Conference on Multimedia Retrieval}{June 30-July 3,
2025}{Chicago, IL, USA}
\acmBooktitle{Proceedings of the 2025 International Conference on Multimedia
Retrieval (ICMR '25), June 30-July 3, 2025, Chicago, IL, USA}
\acmDOI{10.1145/3731715.3733342}
\acmISBN{979-8-4007-1877-9/2025/06}

\begin{document}

\title{FLAIN: Mitigating Backdoor Attacks in Federated Learning via Flipping Weight Updates of Low-Activation Input Neurons}

\author{Binbin Ding}

\affiliation{%
  \institution{Nanjing University of Aeronautics and Astronautics
}
  \city{Nanjing}
  \state{}
  \country{China}
}
\email{dingbinb@nuaa.edu.cn}

\author{Penghui Yang}
\affiliation{%
  \institution{Nanyang Technological University}
  \city{}
  \country{Singapore}}
\email{phyang.cs@gmail.com}

\author{Sheng-Jun Huang}
\authornote{Corresponding author.}
\affiliation{%
  \institution{Nanjing University of Aeronautics and Astronautics
}
  \city{Nanjing}
  \state{}
  \country{China}
}
\email{huangsj@nuaa.edu.cn}

\renewcommand{\shortauthors}{Binbin Ding, Penghui Yang, and Sheng-Jun Huang}

\begin{abstract}
 Federated learning (FL) enables multiple clients to collaboratively train machine learning models under the coordination of a central server, while maintaining privacy. However, the server cannot directly monitor the local training processes, leaving room for malicious clients to introduce backdoors into the model. Research has shown that backdoor attacks exploit specific neurons that are activated only by malicious inputs, remaining dormant with clean data. Building on this insight, we propose a novel defense method called \textbf{F}lipping Weight Updates of \textbf{L}ow-\textbf{A}ctivation \textbf{I}nput \textbf{N}eurons (FLAIN) to counter backdoor attacks in FL. Specifically, upon the completion of global training, we use an auxiliary dataset to identify low-activation input neurons and iteratively flip their associated weight updates. This flipping process continues while progressively raising the threshold for low-activation neurons, until the model's performance on the auxiliary data begins to degrade significantly. Extensive experiments demonstrate that FLAIN effectively reduces the success rate of backdoor attacks across a variety of scenarios, including Non-IID data distributions and high malicious client ratios (MCR), while maintaining minimal impact on the performance of clean data. The source code is available at: \textcolor{blue}{\href{https://github.com/dingbinb/FLAIN}{FLAIN}}.
\end{abstract}



\maketitle

\section{Introduction}
Federated Learning (FL) \cite{c:1}, an emerging distributed machine learning framework, facilitates collaborative model training while preserving the privacy of decentralized data among clients. In recent years, FL has found extensive applications across various domains, including healthcare \cite{c:2,c:3}, financial services \cite{c:4,c:5}, and intelligent transportation \cite{c:6,c:7,c:8}.

In FL, while privacy settings effectively protect participants' data privacy, they also prevent the server coordinating the global training from supervising the underlying processes, leading to a series of new security issues. A prominent concern is that malicious clients can manipulate local training data and model parameters to carry out backdoor attacks \cite{c:10,c:11,c:9,c:12}. A backdoor attack embeds malicious behavior into the machine learning model, allowing it to function normally under standard conditions but to exhibit attacker-controlled behavior when specific triggers appear. As a result, the model maintains normal performance on clean data while producing targeted incorrect results when exposed to the trigger data.

Currently, existing backdoor defense methods can be broadly categorized into two main types: \textit{mid-training defenses} and \textit{post-training defenses}. The former type of methods such as Krum \cite{c:13}, Bulyan \cite{c:14}, Median \cite{c:15} and Trimmed Mean \cite{c:15} mainly seek to identify and filter out malicious updates by scrutinizing discrepancies among the submitted data. These strategies aim to distinguish between benign and potentially malicious updates by detecting anomalies and inconsistencies. However, in scenarios involving Non-Independent and Identically Distributed (Non-IID) data \cite{c:16,c:17,c:33}, the inherent variability in data distribution can amplify discrepancies among benign updates, thereby facilitating the evasion of detection by malicious updates. Recent research \cite{c:18} confirms that benign and malicious updates exhibit a mixed distribution under Non-IID conditions, with anomaly detection based on linear similarity failing to identify backdoors effectively \cite{c:19,c:21,c:20}. RLR \cite{c:28} is different from mainstream mid-training methods, which introduces a robust learning rate adjustment method to counter backdoor attacks, but it is still weakened under Non-IID conditions and its effectiveness is also constrained by the malicious client ratio (MCR). Furthermore, existing research \cite{c:15,c:31,c:29,c:30} highlights the critical need to keep the MCR below 50\% in each training round to ensure the effectiveness of all these mid-training defensive mechanisms.

In post-training defenses, the necessity to identify update discrepancies across clients is eliminated, in contrast to the requirements of mid-training defenses. BadNets \cite{c:22} reveals that trigger samples activate a specific subset of neurons in compromised models, which otherwise remain dormant when exposed to clean inputs. These backdoor neurons often respond exclusively to trigger patterns \cite{c:25} and exhibit high sensitivity to specific inputs \cite{c:26} or perturbations \cite{c:27}. Building on this insight, subsequent studies \cite{c:23,c:24} demonstrate that pruning low-activation neurons effectively mitigates backdoor attacks while preserving overall model performance. Such defenses concentrate on targeting and eliminating neurons with low-activation signals, thereby effectively preventing their activation by malicious inputs. \citet{c:32} applies pruning techniques for backdoor defense in FL and notes that the effectiveness of pruning depends on the attacker’s objectives. When backdoor and benign behaviors rely on the same neurons, pruning these neurons can significantly degrade the model’s performance on both malicious and clean data. The issue arises due to the neurons targeted for removal often playing crucial and essential roles in both backdoor attacks and benign inferences, which considerably reduces the model’s overall performance.

Drawing from the aforementioned research, we attempt to implement backdoor defenses that can eliminate the adverse effects of harmful neurons while preserving the function of normal neurons. In our method, we mainly focus on low-activation signals after they reach the fully connected layer.  For these low-activation inputs, a potential strategy is to prune the associated neurons, preventing the inputs of these neurons from evolving into high-activation signals during backdoor inference. Inspired by RLR \cite{c:28}, which utilizes a negative learning rate to flip the updates of low-confidence dimensions during training, we explore an alternative approach called \textbf{F}lipping Weight Updates of \textbf{L}ow-\textbf{A}ctivation \textbf{I}nput \textbf{N}eurons (FLAIN). Unlike pruning, which removes the output of these neurons, FLAIN flips the direction of the weight updates for these low-activation input neurons obtained during the entire training process and applies these adjusted updates to the initial weights of the fully connected layer. Additionally, to obtain a better flipping interval for low-activation inputs, we introduce a performance-adaptive threshold method that adjusts the flipping interval by a step size each time and evaluates the model's performance variation, until the performance drop reaches a tolerance threshold. This approach ensures that the largest possible flipping interval is achieved for better defense effectiveness, while also maintaining the model's benign performance. Through extensive experiments, we compare FLAIN with the baseline Pruning method. The results demonstrate that FLAIN is more effective at adapting to a broader range of attack scenarios, outperforming pruning low-activation inputs and showing superior defensive robustness.

Our main contributions can be summarized as follows:
\begin{itemize}
\vspace{10pt}
\item We explore a defense strategy for low-activation input neurons in the fully connected layer and propose a method, FLAIN, which mitigates backdoor attacks by flipping the weight updates associated with these neurons obtained during training.
\vspace{10pt}
\item We introduce a performance-adaptive threshold for filtering low-activation inputs, which increases incrementally with each step. As the threshold rises, more neuron weight updates are flipped. This process continues until the performance drop of the updated model becomes unacceptable.

\vspace{4pt}
\item We validate FLAIN's consistent performance improvements across various settings, especially those with Non-IID data distribution and high MCRs. The results demonstrate that FLAIN effectively lowers the success rate of backdoor attacks while preserving the model’s performance on clean data in these settings.
\end{itemize}

\section{Related Work}
\textbf{Backdoor Attacks in FL.} Common strategies for conducting backdoor attacks involve manipulating models to make specific predictions by embedding triggers within samples. These triggers can take various forms: numerical \cite{c:11}, exploiting exact numerical values; semantic \cite{c:34}, influencing predictions through subtle semantic cues; and even imperceptible forms \cite{c:35}, which activate the backdoor despite being hard to detect. \citet{c:36} proposes decomposing a global trigger into localized variants across multiple malicious clients to enhance attack stealth.

\vspace{3pt}
\noindent\textbf{Mid-training  Defenses in FL.} In the quest to combat backdoor attacks, various defense mechanisms emerge. The Krum algorithm \cite{c:13} focuses on selecting the global model by minimizing the sum of Euclidean distances to other local models. Similarly, the Trimmed Mean algorithm \cite{c:15} mitigates the influence of outliers by discarding the highest and lowest values, then calculating the average of the remaining data points. The Bulyan algorithm \cite{c:14} builds on this approach by first using Krum \cite{c:13} to identify a more reliable subset of updates and then applying Trimmed Mean \cite{c:15} for aggregation. RLR \cite{c:28} utilizes robust learning rates to update the global model, applying a negative learning rate to dimensions with significant directional disparities to mitigate the impact of malicious updates. However, while these methods are effective in IID scenarios, their performance significantly diminishes in Non-IID environments due to increased variability among client updates \cite{c:19,c:20}.

\vspace{3pt}
\noindent\textbf{Post-training Defenses in FL.} For a backdoored model, certain neurons are activated by trigger samples but remain dormant with clean inputs. Pruning \cite{c:23,c:24} utilizes clean data to identify and remove these  neurons, minimizing their impact during attacks. \citet{c:32} presents a strategy for FL where clients rank filters based on average activations from local data, and the server aggregates these rankings to prune the global model. Additionally, extreme weights are adjusted after pruning to enhance the robustness of the defense.

\begin{figure*}[tb]
\centering
\includegraphics[width=1.28\columnwidth,height=0.28\textheight]{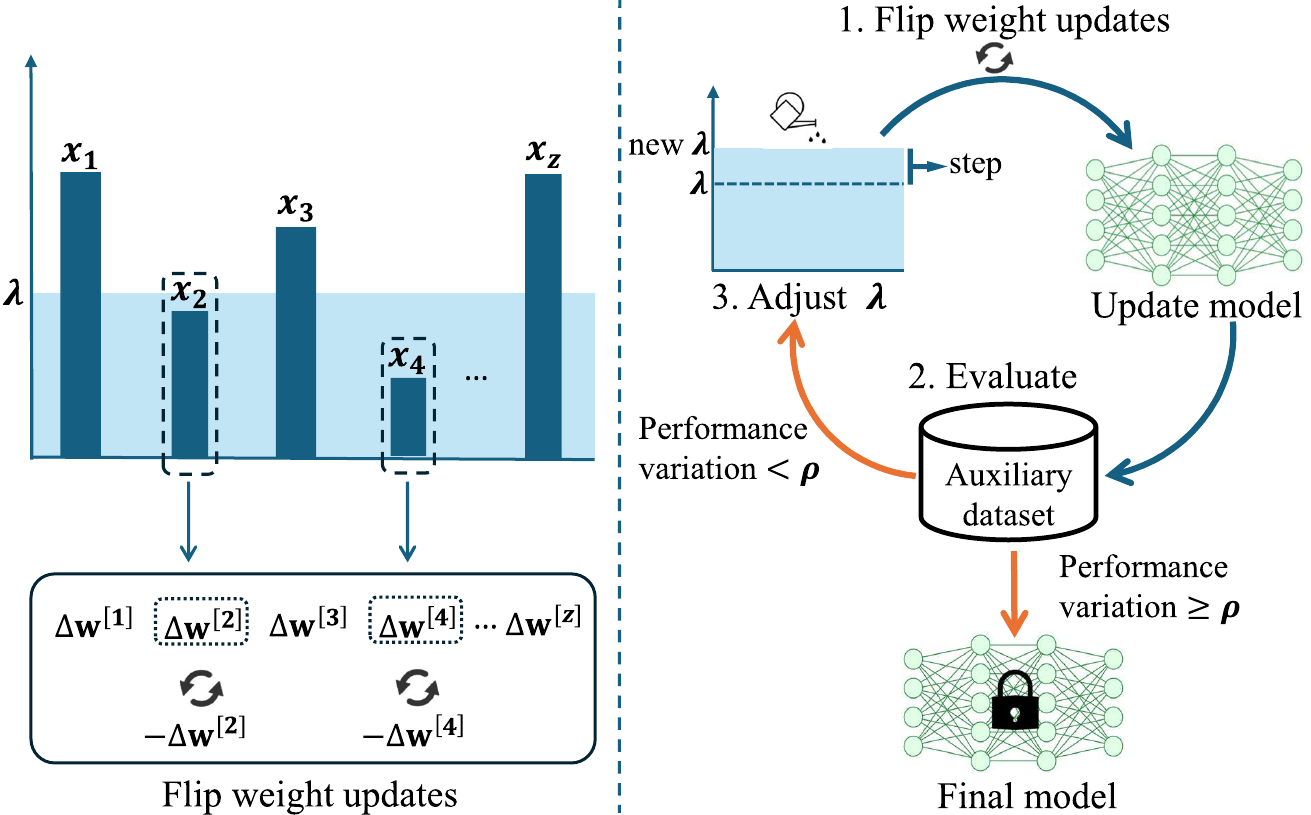} 
\caption{An illustration of FLAIN.} \begin{justify}\textbf{Left} side describes the process of flipping weight updates for neurons with average activation inputs below $\lambda$. 

\noindent\textbf{Right} side introduces the process of using the performance-adaptive method to obtain the optimal threshold $\lambda$, which consists of three steps: \textit{1. Flip weight updates:} The server applies the threshold $\lambda$ to perform the flipping process as outlined on the \textbf{Left} side. \textit{2. Evaluate:} The server evaluates the performance of the updated model and compares the performance variation to the tolerance threshold $\rho$. \textit{3. Adjust $\lambda$:} If the performance variation is smaller than $\rho$, the server increases $\lambda$ incrementally by a defined $step$ size, expanding the flipping interval and re-executing \textit{Flip weight updates}. These steps are repeated until the performance variation exceeds $\rho$, after which the final model is obtained.\end{justify}

\label{fig1:main}
\end{figure*}

\section{Problem Setup}
\subsection{Federated Learning}
Suppose there are $N$  clients in a FL system and the $k$-th client has a local dataset $\mathcal{D}_{k}\!\!=\!\!{\{(x_{k,i},y_{k,i})\}_{i=1}^{n_{k}}}$ of size $n_k$. The server manages client participation in each training round, with the global training objective defined as follows:
\begin{equation}
\mathbf{w} ^{*} = \arg \min_{\mathbf{w} }\textstyle \sum_{k=1}^{N} \lambda_{k} f(\mathbf{w} , \mathcal{D}_{k}),
\end{equation}
\begin{equation}
f(\mathbf{w} , \mathcal{D}_{k}) = \frac{1}{n_{k}} \textstyle \sum_{i=1}^{n_{k}} f(\mathbf{w} , (x_{k,i}, y_{k,i})),
\end{equation}
where $\mathbf{w} ^{*}$ denotes the optimal global model parameters, 
$f(\mathbf{w} , \mathcal{D}_{k})$ represents the average loss computed over the dataset $\mathcal{D}_k$ for client $k$,
$(x_{k,i}, y_{k,i})$ denotes the $i$-th sample in 
$\mathcal{D}_k$, and $\lambda_{k}$ indicates the weight of the loss for client 
$k$.

At the $t$-th federated training round, the server randomly selects a client set $S_t$ where $|S_t|=m$ and $0<m\le N$, and broadcasts the current model parameters $\mathbf{w}_t$. The selected clients perform local training and upload their updates $\bigtriangleup \mathbf{w}_{t+1}^{\,k}$ to the server for global model aggregation. FedAvg \cite{c:1} uses a weighted averaging rule for aggregating client updates according to the volume of local data. The aggregation process is detailed as follows:
\begin{equation}
\mathbf{w}_{t+1}= \mathbf{w}_{t} + \alpha \cdot\frac{ {\textstyle \sum_{k \,\in\, S_t}n_{k}\bigtriangleup \mathbf{w}_{t+1}^{\,k}  } }{ {\textstyle \sum_{k \,\in \, S_t}n_{k}} }, 
\end{equation}
where $\alpha$ denotes the global learning rate and  $\bigtriangleup \mathbf{w}_{t+1}^{\,k}$ represents the local training update from client $k$. 
\subsection{Threat Model}
Due to privacy constraints, the server lacks access to local training data and processes. 
\citet{c:11} reveals vulnerabilities in FedAvg \cite{c:1}, highlighting its susceptibility to backdoor attacks.

\vspace{3pt}
\noindent\textbf{Attacker’s Capabilities.} We assume that malicious attackers have access to the local training data of compromised clients and are able to modify the samples of targeted classes. However, they are unable to access the data or model parameters of non-compromised clients, nor can they disrupt the server's aggregation process.

\vspace{3pt}
\noindent\textbf{Attacker’s Goals.} 
The backdoored model is engineered to predict a specific class for samples embedded with triggers.  For any input $x$, the ideal output of the backdoored model $\mathbf{M_{w}}$ can be formulated as follows:
\begin{equation}
 \mathbf{M_{w}}(x) = \begin{cases}\,y & \text{if } x \in \mathcal{D}_{clean}, \\\,y_{target} & \text{else},\end{cases}
 \end{equation}
and the training objective can be formulated as follows:
\begin{equation}
\begin{aligned}
\min_{\mathbf{w}} \ & \underbrace{\mathbb{E}_{(x, \,y) \sim \mathcal{D}_{clean}} \, \ell(\mathbf{M_{w}}(x), \,y)}_{\text{Average loss on clean data}} \\
& + \lambda \cdot \underbrace{\mathbb{E}_{(x', \,\,y_{target}) \sim \mathcal{D}_{poison}} \, \ell(\mathbf{M_{w}}(x'), 
 \,y_{target})}_{\text{Average loss on poisoned data}},
\end{aligned}
\end{equation}
where \( \ell \) is the loss function, $y_{target}$ is the target label,  \( \mathcal{D}_{clean} \) represents the clean dataset, \( \mathcal{D}_{\mathit{poison}} \) represents the poisoned dataset, and \( \lambda \) is a balancing parameter. 
\subsection{Defense Model}
\noindent\textbf{Defender’s Capabilities.} 
Consistent with prior research \cite{c:40,c:39,c:37,c:38}, we assume that the server maintains a small auxiliary dataset, which is used to compute the activation inputs of the fully connected layer and to evaluate performance variations within the model.

\vspace{3pt}
\noindent\textbf{Defender’s\,\, Goals.}
The backdoored model integrates information learned from both clean and trigger data, making it challenging to selectively remove backdoor information. Therefore, the defender's objectives are twofold: 1) curtailing the model's accuracy on trigger samples; 2) preserving its performance on clean data.

\section{Methodology}

In this section, we explain in detail the process of handling low-activation input neurons in the fully connected layer using \textbf{Flipping} and the \textbf{Performance-Adaptive Threshold}, which together constitute the FLAIN. 

\subsection{Flipping}\label{sec:flipping}

\noindent\textbf{Activated Inputs.} Consider a model that includes a fully connected layer $\tau$ with $z$ neurons, where the weight matrix $\mathbf{W}$ is defined as $\mathbf{W}=\left(\mathbf{w}^{[1]}, \mathbf{w}^{[2]}, \cdots, \mathbf{w}^{[z]}\right) \in \mathbb{R}^{s \times z}$. Each column vector $\mathbf{w}^{[i]} = (w^{[i]_{1}}, w^{[i]_{2}},\cdots, w^{[i]_{s}})^{\top}\in\mathbb{R}^{s}$ in 
 $\mathbf{W}$ represents the weights associated with the $i$-th neuron in layer $\tau$, where $w^{[i]_{j}}(j=1,2,\cdots,s)$ denotes the $j$-th element of the vector $\mathbf{w}^{[i]}$. Let $\mathbf{x} =({x}_1, {x} _2,\cdots,{x}_z)^{\top} \in \mathbb{R}^{z}$ denote the inputs to layer $\tau$. These inputs are obtained by applying the $\mathrm{ReLU}$ activation function \cite{c:41} element-wise to the raw outputs from the preceding layer. Specifically, each component $x_i (i=1,2,\cdots,z)$ of the vector $\mathbf{x}$ is defined as $x_i=\mathrm{ReLU}(x'_i)$, where $x'_i$ represents the raw output from the preceding layer. The output $\mathcal{H}(\mathbf{x} ) $ of layer $\tau$ can be expressed as:
\begin{equation}
\begin{aligned}
\mathcal{H}(\mathbf{x} ) &= \mathbf{W} \mathbf{x}  + \mathbf{b}  \\
     &= \begin{bmatrix}w^{[1]_{1}}x_{1} + w^{[2]_{1}}x_{2} +\cdots+ w^{[z]_{1}}x_{z}+ b _1 \\w^{[1]_{2}}x_{1} + w^{[2]_{2}}x_{2} +\cdots+ w^{[z]_{2}}x_{z}+ b_2\\
     \cdots\\ 
     w^{[1]_{s}}x_{1} + w^{[2]_{s}}x_{2} +\cdots+ w^{[z]_{s}}x_{z}+ b_s \\\end{bmatrix}
\end{aligned},
\end{equation}
where $\mathbf{b} = (b_1, b_2, \cdots, b_s)^{\top} \in \mathbb{R}^{s}$ is the bias vector for layer $\tau$.

\vspace{3pt}
\noindent\textbf{Flipping.} FLAIN offers a new approach for handling low-activation input neurons. In contrast to pruning that sets the weights of these neurons to 0
, FLAIN modifies these weights by flipping the updates obtained during global training. Before training begins, the server initializes the global model parameters and records the initial weights of layer $\tau$:  $\mathbf{W}_{0}=\left(\mathbf{w}_{0}^{[1]}, \mathbf{w}_{0}^{[2]}, \cdots, \mathbf{w}_{0}^{[z]}\right) \in \mathbb{R}^{s \times z}$. 

Upon completing the entire global training, the server acquires layer 
$\tau$'s weights: $\mathbf{W}\!=\!\left(\mathbf{w}^{[1]}, \mathbf{w}^{[2]}, \ldots, \mathbf{w}^{[z]}\right) \!\in \mathbb{R}^{s \times z}$ and computes the weight updates: $\Delta \mathbf{W }\! = \!\mathbf{W }\!-\! \mathbf{W} _{0}\!=\!(\Delta \mathbf{w}^{[1]},\Delta \mathbf{w }^{[2]}, \ldots, \Delta \mathbf{w }^{[z]}) \!\in\! \mathbb{R}^{s \times z}$. Using an auxiliary dataset $\psi$ consisting of clean data, with the assumption that the sample size for each class is balanced, the server simultaneously evaluates the model's performance, denoted as $acc_0$, and computes the activation input values for layer $\tau$. Specifically, for each sample in the auxiliary dataset, the server performs forward propagation and records the activation input values for each neuron in layer $\tau$. The recorded activation input values for each neuron are then averaged across all samples in the dataset. This results in the activation input vector $\mathbf{x} = ({x}_1, {x}_2, \cdots, {x}_z)^{\top} \in \mathbb{R}^z$, where each $x_i$ represents the average activation input for the $i$-th neuron in layer $\tau$. The server then sets a threshold $\lambda$, and all $x_i$ values that are not greater than $\lambda$ are collected into the set $\mathcal{D}_{flip}$. The detailed process for selecting the optimal threshold $\lambda$ is described in Subsection \textbf{~\ref{sec:Performance-Adaptive Threshold}}. Finally, the server flips the weight updates $\Delta \mathbf{w}^{[i]}$ to $-\Delta \mathbf{w}^{[i]}$ for the neurons in $\mathcal{D}_{flip}$, and uses these flipped updates to update $\mathbf{W}_0$, as formulated below:

\begin{equation}
  \mathbf{w}^{[i]^*}=\begin{cases} \mathbf{w}_{0}^{[i]} - \Delta \mathbf{w} ^{[i]}, &\text{if}\,\, x_i \in \mathcal{D}_{flip}, \\ \mathbf{w}_{0}^{[i]} + \Delta\mathbf{w} ^{[i]}, &\text{else}.\end{cases}
\end{equation}

The whole process of flipping is illustrated on the \textbf{Left} side of Figure \textbf{\ref{fig1:main}}.

\begin{algorithm}[tb]
\caption{Flipping Weight Updates of Low-Activation Input Neurons (FLAIN)}
\label{alg:algorithm1}
\begin{justify}
    
\textbf{Input}: the fully connected layer $\tau$'s initial weights $\mathbf{W}_{0}$ and the number of neurons $z$; model parameters $\theta$ and layer $\tau$'s weights $\mathbf{W}$ at the end of training; the $step$ size; performance variation threshold $\rho$; an auxiliary dataset $\psi$.
\end{justify}
\begin{algorithmic}[1] 
\State $n_{0} = ||\mathbf{W}||_{2}$
\State $\Delta \mathbf{W } = \mathbf{W } - \mathbf{W} _{0}$
\State $ \mu,\,\mathbf{x} = \mathrm{Extract}(\theta, \psi)$
\State $ \lambda = \mu + step $
\State $acc_{0} = \mathrm{Evaluate}(\theta, \psi)$
\While{True}
\For{$i = 1, 2, 3, \cdots, z$}
\If{$x_i \le \lambda$}
\State $\mathbf{w}^{[i]^*} = \mathbf{w}_{0}^{[i]} - \Delta \mathbf{w} ^{[i]}$
\Else
\State $\mathbf{w}^{[i]^*} = \mathbf{w}_{0}^{[i]} + \Delta \mathbf{w} ^{[i]}$
\EndIf
\EndFor
\State  $\mathbf{W}^*=\left(\mathbf{w}^{[1]^*}, \mathbf{w}^{[2]^*}, \cdots, \mathbf{w}^{[z]^*}\right)$
\State  $\theta^* = \mathrm{Update}(\theta,  \mathbf{W}, \mathbf{W}^*)$
\State $acc_{1} = \mathrm{Evaluate}(\theta^*, \psi)$
\If {$\rho \le acc_{0} - acc_{1}$}
\State $n_{1} = ||\mathbf{W}^*||_{2}$
\State $\mathbf{W}^* = \mathbf{W}^* \times \frac{n_{0}}{n_{1}}$
\State  $\theta^* = \mathrm{Update}(\theta,  \mathbf{W}, \mathbf{W}^*)$
\State \Return $\theta^{*}$ 
\Else
\State $\lambda = \lambda + step$
\EndIf
\EndWhile
\end{algorithmic}
\begin{justify}
\textbf{Output}: final model parameters $\theta^{*}$.
\end{justify}
\end{algorithm}

\subsection{Performance-Adaptive Threshold} \label{sec:Performance-Adaptive Threshold}

The threshold $\lambda$ proposed in Subsection \textbf{~\ref{sec:flipping}} significantly determines the tradeoff between clearing backdoors and maintaining performance on clean data. To automatically identify a better $\lambda$, we propose a performance-adaptive method for threshold selection.

When first evaluating the fully-trained model using the auxiliary dataset, the server obtains the average activation input vector $\mathbf{x} = ({x}_1, {x}_2, \cdots, {x}_z)^{\top} \in \mathbb{R}^{z}$ for layer $\tau$ across the entire auxiliary dataset and identifies the minimum average activation value $\mu (\mu \!\in\! \mathbf{x})$. Then the server sets an initial threshold $\lambda = \mu + step$, which is slightly larger than $\mu$ with a $step$ size, and performs the flipping process as independently described in Subsection \textbf{~\ref{sec:flipping}}. After updating the weights of layer $\tau$ in the model from $\mathbf{W}$ to $\mathbf{W}^*=\left(\mathbf{w}^{[1]^*}, \mathbf{w}^{[2]^*}, \cdots, \mathbf{w}^{[z]^*}\right) \in \mathbb{R}^{s \times z}$, the server assesses the current model's performance $acc_1$ on the auxiliary dataset $\psi$. 
If the performance gap $\epsilon = acc_{0}-acc_{1}$ is smaller than the tolerance threshold 
$\rho$, the server increments $\lambda$ by a $step$ size, which results in the identification of more neurons with low activation. This causes the associated weight updates to be flipped within a larger range, thereby flipping more column vectors of the update matrix $\Delta \mathbf{W}$. The procedure is repeated until 
$\epsilon$ exceeds the threshold 
$\rho$, indicating that the performance drop has surpassed the server’s tolerance limit. This adaptive threshold modification process is illustrated on the \textbf{Right} side of Figure \textbf{\ref{fig1:main}}.

The details of the whole FLAIN method are summarized in Figure \textbf{\ref{fig1:main}} and the pseudo codes are given in Algorithm \textbf{\ref{alg:algorithm1}}.

\begin{table*}[]
\caption{Defense performance under IID data.  \textuparrow\ indicates larger values are better, and \textdownarrow\ indicates smaller values are better. \textbf{Bold} numbers are superior results.}
\vspace{-5pt}
\setlength{\tabcolsep}{1.95mm}
\centering
\begin{tabular}{|ccc|ccc|ccc|ccc|ccc|}
\hline
\multicolumn{3}{|c|}{\textbf{Dataset}}                              & \multicolumn{3}{c|}{MNIST}                                   & \multicolumn{3}{c|}{FMNIST}                            & \multicolumn{3}{c|}{EMNIST}                                                       & \multicolumn{3}{c|}{CIFAR-10}                                \\ \hline
\multicolumn{2}{|c|}{\textbf{Attack}}                     & \textbf{Defense} & \multicolumn{1}{c|}{ASR  \textdownarrow} & \multicolumn{1}{c|}{ACC  \textuparrow} & OPS  \textuparrow    & \multicolumn{1}{c|}{ASR \textdownarrow} & \multicolumn{1}{c|}{ACC  \textuparrow} & OPS  \textuparrow   & \multicolumn{1}{c|}{ASR \textdownarrow} & \multicolumn{1}{c|}{ACC  \textuparrow} & \multicolumn{1}{c|}{OPS  \textuparrow}    & \multicolumn{1}{c|}{ASR \textdownarrow} & \multicolumn{1}{c|}{ACC  \textuparrow} & OPS  \textuparrow    \\ \hline
\multicolumn{1}{|c|}{}  & \multicolumn{1}{c|}{}  & FedAvg  & 1                        & 0.991                    & 0      & 0.990                    & 0.918                    & 0      & 0.999                    & 0.854                    & \multicolumn{1}{c|}{0}      & 0.973                    & 0.856                    & 0      \\\cline{3-15}
\multicolumn{1}{|c|}{}  & \multicolumn{1}{c|}{}  & Krum    & 0.367                    & 0.990                    & +0.632  & 0.992                    & 0.897                    & -0.025 & 0.913                    & 0.842                    & \multicolumn{1}{c|}{+0.072}  & 0.851                    & 0.778                    & +0.034  \\
\multicolumn{1}{|c|}{}  & \multicolumn{1}{c|}{}  & Median  & 1                        & \textbf{0.991}                    & 0      & 0.991                    & \textbf{0.917}                    & -0.002 & 0.977                    & 0.844                    & \multicolumn{1}{c|}{+0.010}   & 0.973                    & 0.843                    & -0.015 \\
\multicolumn{1}{|c|}{}  & \multicolumn{1}{c|}{5} & RLR     & \textbf{0}                        & 0.980                    & +0.989  & 0.004                    & 0.874                    & +0.948  & 0.013                    & 0.838                    & \multicolumn{1}{c|}{+0.968}  & 0.971                    & 0.839                    & -0.018 \\
\multicolumn{1}{|c|}{}  & \multicolumn{1}{c|}{}  & FLTrust & 0.993                    & 0.985                    & +0.001  & 0.512                    & 0.871                    & +0.432  & 0.047                    & 0.762                    & \multicolumn{1}{c|}{+0.845}  & 0.915                    & 0.753                    & -0.061 \\
\multicolumn{1}{|c|}{}  & \multicolumn{1}{c|}{}  & Pruning & 0.001                    & \textbf{0.991}                    & +0.999  & 0.002                    & 0.906                    & +0.985  & \textbf{0}                        & \textbf{0.854}                    & \multicolumn{1}{c|}{\textbf{+1}}      & 0.841                    & 0.850                    & +0.129  \\
\multicolumn{1}{|c|}{}  & \multicolumn{1}{c|}{}  & M-metrics & \textbf{0}                    & \textbf{0.991}                   & \textbf{+1}  & 0.986                    & 0.906                    & -0.009  & 0.013                    & 0.817                   & \multicolumn{1}{c|}{+0.944}  & 0.942                    & 0.845                    & +0.019  \\
\multicolumn{1}{|c|}{}  & \multicolumn{1}{c|}{}  & FLAIN   & \textbf{0}                        & \textbf{0.991}                    & \textbf{+1}      & \textbf{0}                        & 0.908                    & \textbf{+0.989}  & \textbf{0}                        & \textbf{0.854}                    & \multicolumn{1}{c|}{\textbf{+1}}      & \textbf{0.001}                    & \textbf{0.851}                    & \textbf{+0.993}  \\ \cline{2-15} 
\multicolumn{1}{|c|}{}  & \multicolumn{1}{c|}{}  & FedAvg  & 1                        & 0.992                    & 0      & 0.994                    & 0.919                    & 0      & 1                        & 0.853                    & \multicolumn{1}{c|}{0}      & 0.974                    & 0.843                    & 0      \\\cline{3-15}
\multicolumn{1}{|c|}{}  & \multicolumn{1}{c|}{}  & Krum    & 0.709                    & 0.990                    & +0.289  & 0.848                    & 0.898                    & +0.124  & 0.946                    & 0.843                    & +0.042                       & 0.967                    & 0.757                    & -0.095 \\
\multicolumn{1}{|c|}{}  & \multicolumn{1}{c|}{}  & Median  & 1                        & \textbf{0.992}                    & 0      & 0.989                    & 0.902                    & -0.013 & 1                        & 0.851                    & \multicolumn{1}{c|}{-0.002} & 0.973                    & 0.832                    & -0.012 \\
\multicolumn{1}{|c|}{0} & \multicolumn{1}{c|}{6} & RLR     & \textbf{0}                        & 0.978                    & +0.986  & 0.108                    & 0.873                    & +0.841  & 0.024                    & 0.834                    & \multicolumn{1}{c|}{+0.954}  & 0.964                    & 0.793                    & -0.049 \\
\multicolumn{1}{|c|}{}  & \multicolumn{1}{c|}{}  & FLTrust & 1                        & 0.986                    & -0.006 & 0.573                    & 0.881                    & +0.382  & 0.179                    & 0.742                    & \multicolumn{1}{c|}{+0.691}  & 0.882                    & 0.736                    & -0.032 \\
\multicolumn{1}{|c|}{}  & \multicolumn{1}{c|}{}  & Pruning & 0.002                    & 0.987                    & +0.993  & 0.544                    & 0.906                    & +0.439  & 0.003                    & \textbf{0.852}                    & \multicolumn{1}{c|}{+0.996}  & 0.145                    & 0.829                    & +0.835  \\
\multicolumn{1}{|c|}{}  & \multicolumn{1}{c|}{}  & M-metrics & 0.003                    & 0.991                    & +0.996  & 0.977                    & 0.908                    & +0.005  & 0.996                    & 0.836                    & \multicolumn{1}{c|}{-0.016}  & 0.971                   & \textbf{0.845}                    & +0.005  \\
\multicolumn{1}{|c|}{}  & \multicolumn{1}{c|}{}  & FLAIN   & \textbf{0}                        & 0.991                    & \textbf{+0.999}  & \textbf{0.006}                    & \textbf{0.911}                    & \textbf{+0.985}  & \textbf{0}                        & \textbf{0.852}                    & \multicolumn{1}{c|}{\textbf{+0.999}}  & \textbf{0}                        & 0.832                    & \textbf{+0.987}  \\ \cline{2-15} 
\multicolumn{1}{|c|}{}  & \multicolumn{1}{c|}{}  & FedAvg  & 1                        & 0.992                    & 0      & 0.992                    & 0.917                    & 0      & 0.999                    & 0.864                    & \multicolumn{1}{c|}{0}      & 0.971                    & 0.854                    & 0      \\\cline{3-15}
\multicolumn{1}{|c|}{}  & \multicolumn{1}{c|}{}  & Krum    & 0.483                    & 0.990                    & +0.515  & 0.989                    & 0.899                    & -0.017 & 0.933                    & 0.843                    & \multicolumn{1}{c|}{+0.042}  & 0.932                    & 0.770                    & -0.058 \\
\multicolumn{1}{|c|}{}  & \multicolumn{1}{c|}{}  & Median  & 1                        & 0.991                    & -0.001 & 0.979                    & 0.896                    & -0.010  & 0.997                    & 0.847                    & \multicolumn{1}{c|}{-0.018} & 0.971                    & 0.824                    & -0.035 \\
\multicolumn{1}{|c|}{}  & \multicolumn{1}{c|}{7} & RLR     & \textbf{0}                        & 0.983                    & +0.991  & 0.003                    & 0.872                    & +0.948  & 0.027                    & 0.835                    & \multicolumn{1}{c|}{+0.939}  & 0.637                    & 0.713                    & +0.179  \\
\multicolumn{1}{|c|}{}  & \multicolumn{1}{c|}{}  & FLTrust & 1                        & 0.984                    & -0.008 & 0.355                    & 0.875                    & +0.596  & 0.824                    & 0.815                    & \multicolumn{1}{c|}{+0.118}  & 0.948                    & 0.723                    & -0.130  \\
\multicolumn{1}{|c|}{}  & \multicolumn{1}{c|}{}  & Pruning & \textbf{0}                        & \textbf{0.992}                    & \textbf{+1}      & 0.038                    & 0.905                    & +0.949  & \textbf{0}                        & 0.862                    & \multicolumn{1}{c|}{+0.998}  & 0.653                    & 0.812                    & +0.278  \\
\multicolumn{1}{|c|}{}  & \multicolumn{1}{c|}{}  & M-metrics & \textbf{0}                    & 0.990                   & +0.998  & 0.983                    & \textbf{0.908}                    & -0.001  & 0.034                    & 0.814                    & \multicolumn{1}{c|}{+0.908}  & 0.976                   & \textbf{0.841}                    & -0.020  \\
\multicolumn{1}{|c|}{}  & \multicolumn{1}{c|}{}  & FLAIN   & \textbf{0}                        & 0.987                    & +0.995  & \textbf{0}                        & 0.905                    & \textbf{+0.987}  & \textbf{0}                        & \textbf{0.863}                    & \multicolumn{1}{c|}{\textbf{+0.999}}  & \textbf{0.003}                    & 0.828                    & \textbf{+0.966}  \\ \hline
\end{tabular}
\label{tab:table1}
\end{table*}

\section{Performance \,Evaluation}
\subsection{Experimental  Setup} 
\noindent\textbf{Training Settings.} For our experiments, we use the \textit{Adam} optimizer \cite{c:42} with a learning rate of 0.001 and betas set to (0.9, 0.999). Each training round consists of local training for 1 epoch with a batch size of 256, while both the client sampling rate and the global learning rate are set to 1. 

\vspace{3pt}
\noindent\textbf{Evaluation Metrics.} We utilize the following three metrics to assess the model's performance.
\begin{itemize}
\item \noindent\textbf{Attack Success Rate (ASR).} ASR measures the proportion of trigger samples that the model correctly classifies as the target class. This metric indicates the effectiveness of the backdoor attack, reflecting how well the attack can manipulate the model to produce the desired output for malicious inputs.
\item\noindent\textbf{Natural Accuracy (ACC).} ACC quantifies the model's performance on untainted data, representing its effectiveness in standard classification tasks. This metric reflects how well the model generalizes to non-malicious inputs.
\vspace{3pt}
\item \textbf{Overall Performance Score (OPS).} Following \cite{c:48}, we use the OPS score to evaluate the performance improvements of different defense methods in terms of ACC and ASR relative to the baseline FedAvg. The OPS is calculated as 
\( \mathrm{OPS} = \frac{D_{\mathrm{ACC}} - B_{\mathrm{ACC}}}{B_{\mathrm{ACC}}} - \frac{D_{\mathrm{ASR}} - B_{\mathrm{ASR}}}{B_{\mathrm{ASR}}} \), where 
\(D_{\mathrm{ACC}}\) and \(D_{\mathrm{ASR}}\)  represent the ACC and ASR of the current defense method, and 
\(B_{\mathrm{ACC}}\) and \(B_{\mathrm{ASR}}\) are the baseline FedAvg's values.
\end{itemize}

\begin{table*}[]
\caption{Defense performance under Non-IID data.}
\vspace{-5pt}
\setlength{\tabcolsep}{1.95mm}
\centering
\begin{tabular}{|ccc|ccc|ccc|ccc|ccc|}
\hline
\multicolumn{3}{|c|}{\textbf{Dataset}}                              & \multicolumn{3}{c|}{MNIST}                                                     & \multicolumn{3}{c|}{FMNIST}                                              & \multicolumn{3}{c|}{EMNIST}                                                    & \multicolumn{3}{c|}{CIFAR-10}                                                  \\ \hline
\multicolumn{2}{|c|}{\textbf{Attack}}                     & \textbf{Defense} & \multicolumn{1}{c|}{ASR \textdownarrow} & \multicolumn{1}{c|}{ACC \textuparrow} & \multicolumn{1}{l|}{OPS \textuparrow} & \multicolumn{1}{c|}{ASR \textdownarrow} & \multicolumn{1}{c|}{ACC \textuparrow} & \multicolumn{1}{l|}{OPS \textuparrow} & \multicolumn{1}{c|}{ASR \textdownarrow} & \multicolumn{1}{c|}{ACC \textuparrow} & \multicolumn{1}{l|}{OPS \textuparrow} & \multicolumn{1}{c|}{ASR \textdownarrow} & \multicolumn{1}{c|}{ACC \textuparrow} & \multicolumn{1}{l|}{OPS \textuparrow} \\ \hline
\multicolumn{1}{|c|}{}  & \multicolumn{1}{c|}{}  & FedAvg  & 1                        & 0.991                    & 0                        & 0.991                    & 0.911                    & 0                        & 0.999                    & 0.846                    & 0                        & 0.919                    & 0.777                    & 0                        \\\cline{3-15}
\multicolumn{1}{|c|}{}  & \multicolumn{1}{c|}{}  & Krum    & 0.893                    & 0.971                    & +0.087                    & 0.549                    & 0.808                    & +0.333                    & \textbf{0}                        & 0.706                    & +0.835                    & 0.133                    & 0.332                    & +0.283                    \\
\multicolumn{1}{|c|}{}  & \multicolumn{1}{c|}{}  & Median  & 0.770                    & 0.969                    & +0.208                    & 0.989                    & \textbf{0.909}                    & 0                        & 0.998                    & 0.840                     & -0.006                   & 0.953                    & 0.737                    & -0.088                   \\
\multicolumn{1}{|c|}{}  & \multicolumn{1}{c|}{5} & RLR     & \textbf{0}                        & 0.966                    & +0.975                    & \textbf{0.001}                    & 0.762                    & +0.835                    & \textbf{0}                        & 0.805                    & +0.952                    & 0.791                    & 0.663                    & -0.007                   \\
\multicolumn{1}{|c|}{}  & \multicolumn{1}{c|}{}  & FLTrust & 0.959                    & 0.983                    & +0.033                    & 0.767                    & 0.860                     & +0.170                     & 0.026                    & 0.744                    & +0.853                    & 0.734                    & 0.643                    & +0.029                    \\
\multicolumn{1}{|c|}{}  & \multicolumn{1}{c|}{}  & Pruning & 0.001                    & 0.986                    & +0.994                    & 0.560                     & 0.901                    & +0.424                    & \textbf{0}                        & \textbf{0.844}                    & \textbf{+0.998}                    & 0.792                    & 0.732                    & +0.080                     \\
\multicolumn{1}{|c|}{}  & \multicolumn{1}{c|}{}  & M-metrics & 0.999                    & 0.983                    & -0.007  & 0.792                    & 0.893                   & +0.181  & 0.008                    & 0.830                    & \multicolumn{1}{c|}{+0.973}  & 0.051                    & 0.724                    & +0.876  \\

\multicolumn{1}{|c|}{}  & \multicolumn{1}{c|}{}  & FLAIN   & \textbf{0}                        & \textbf{0.987}                    & \textbf{+0.996}                    & 0.005                    & 0.903                    & \textbf{+0.986}                    & \textbf{0}                        & \textbf{0.844}                    & \textbf{+0.998}                    & \textbf{0.019}                    & \textbf{0.751}                    & \textbf{+0.946}                    \\ \cline{2-15} 
\multicolumn{1}{|c|}{}  & \multicolumn{1}{c|}{}  & FedAvg  & 1                        & 0.991                    & 0                        & 0.989                    & 0.905                    & 0                        & 0.999                    & 0.846                    & 0                        & 0.943                    & 0.794                    & 0                        \\\cline{3-15}
\multicolumn{1}{|c|}{}  & \multicolumn{1}{c|}{}  & Krum    & 0.987                    & 0.972                    & -0.006                   & 0.877                    & 0.813                    & +0.007                    & \textbf{0}                        & 0.721                    & +0.852                    & 0.041                    & 0.342                    & +0.396                    \\
\multicolumn{1}{|c|}{}  & \multicolumn{1}{c|}{}  & Median  & 0.999                    & 0.982                    & -0.008                   & 0.987                    & 0.897                    & -0.011                   & 0.998                    & 0.837                    & -0.010                    & 0.883                    & 0.726                    & -0.026                   \\
\multicolumn{1}{|c|}{0} & \multicolumn{1}{c|}{6} & RLR     & 0.013                    & 0.964                    & +0.960                     & \textbf{0}                        & 0.741                    & +0.813                    & 0.005                    & 0.797                    & +0.937                    & 0.945                    & \textbf{0.769}                    & -0.039                   \\
\multicolumn{1}{|c|}{}  & \multicolumn{1}{c|}{}  & FLTrust & 0.968                    & 0.983                    & +0.024                    & 0.727                    & 0.862                    & +0.213                    & 0.163                    & 0.755                    & +0.729                    & 0.794                    & 0.686                    & +0.019                    \\
\multicolumn{1}{|c|}{}  & \multicolumn{1}{c|}{}  & Pruning & 0.002                    & 0.985                    & +0.992                    & 0.685                    & 0.889                    & +0.285                    & 0.030                     & \textbf{0.845}                    & +0.969                    & 0.886                    & 0.724                    & -0.032                   \\
\multicolumn{1}{|c|}{}  & \multicolumn{1}{c|}{}  & M-metrics & 0.008                    & 0.982                    & +0.983  & 0.234                    & 0.889                    & +0.746  & 0.033                    & 0.807                    & \multicolumn{1}{c|}{+0.921}  & 0.098                    & 0.670                  & +0.740  \\
\multicolumn{1}{|c|}{}  & \multicolumn{1}{c|}{}  & FLAIN   & \textbf{0}                        & \textbf{0.988}                    & \textbf{+0.997}                    & 0.036                    & \textbf{0.900}                      & \textbf{+0.952}                    & \textbf{0}                        & \textbf{0.845}                    & \textbf{+0.999}                    & \textbf{0.018}                    & 0.746                    & \textbf{+0.941}                    \\ \cline{2-15} 
\multicolumn{1}{|c|}{}  & \multicolumn{1}{c|}{}  & FedAvg  & 1                        & 0.990                     & 0                        & 0.977                    & 0.905                    & 0                        & 0.997                    & 0.847                    & 0                        & 0.974                    & 0.801                    & 0                        \\\cline{3-15}
\multicolumn{1}{|c|}{}  & \multicolumn{1}{c|}{}  & Krum    & 0.997                    & 0.975                    & -0.013                   & 0.598                    & 0.808                    & +0.284                    & \textbf{0}                        & 0.726                    & +0.858                    & 0.246                    & 0.372                    & +0.211                    \\
\multicolumn{1}{|c|}{}  & \multicolumn{1}{c|}{}  & Median  & 0.984                    & 0.980                     & +0.005                    & 0.974                    & 0.881                    & -0.016                   & 0.993                    & 0.840                     & -0.001                   & 0.965                    & 0.736                    & -0.103                   \\
\multicolumn{1}{|c|}{}  & \multicolumn{1}{c|}{7} & RLR     & 0.037                    & 0.965                    & +0.937                    & \textbf{0}                        & 0.743                    & +0.816                    & \textbf{0}                        & 0.812                    & +0.960                     & 0.886                    & 0.661                    & -0.113                   \\
\multicolumn{1}{|c|}{}  & \multicolumn{1}{c|}{}  & FLTrust & 0.745                    & 0.982                    & +0.246                    & 0.603                    & 0.865                    & +0.341                    & 0.005                    & 0.755                    & +0.887                    & 0.702                    & 0.670                     & +0.098                    \\
\multicolumn{1}{|c|}{}  & \multicolumn{1}{c|}{}  & Pruning & 0.001                    & 0.990                     & +0.998                    & 0.770                     & 0.894                    & +0.204                    & 0.004                    & \textbf{0.846}                    & +0.996                    & 0.922                    & 0.752                    & -0.035                   \\
\multicolumn{1}{|c|}{}  & \multicolumn{1}{c|}{}  & M-metrics & 0.355                    & 0.984                    & +0.639  & 0.196                    & 0.884                    & +0.776  & 0.027                  & 0.831                    & \multicolumn{1}{c|}{+0.954}  & 0.412                    & 0.703                    & +0.455  \\
\multicolumn{1}{|c|}{}  & \multicolumn{1}{c|}{}  & FLAIN   & \textbf{0}                        & \textbf{0.991}                    & \textbf{+1}                        & 0.002                    & \textbf{0.896}                    & \textbf{+0.982}                    & \textbf{0}                        & 0.844                    & \textbf{+0.998}                    & \textbf{0.017}                    & \textbf{0.784}                    &  \textbf{+0.991}                    \\ \hline
\end{tabular}
\label{tab:table2}
\vspace{5pt}
\end{table*}

\noindent\textbf{Datasets and Models.} 
In our experiments, we utilize a diverse range of well-known datasets, including MNIST \cite{c:43}, FashionMNIST \cite{c:44}, EMNIST \cite{c:45}, and CIFAR-10 \cite{c:46}.
\begin{itemize}
\item \textbf{MNIST} is a highly recognized and extensively utilized dataset consisting of handwritten digits (0-9), with 60,000 training images and 10,000 test images. Each image is 28$\times$28 pixels in grayscale. In our experiments with MNIST, we adopt a CNN architecture featuring 2 convolutional layers followed by 2 fully connected layers.
\vspace{3pt}
\item \textbf{FashionMNIST (FMNIST)} is a popular alternative to MNIST, containing grayscale images from 10 distinct categories of clothing items. The dataset consists of 60,000 training images and 10,000 test images, each with dimensions of 28$\times$28 pixels. For experiments involving FashionMNIST, we utilize a CNN architecture with 2 convolutional layers, followed by 4 fully connected layers.
\vspace{3pt}
\item \textbf{EMNIST} extends MNIST by including handwritten letters alongside digits, offering a broader and more complex dataset. Specifically, we utilize the “byClass” partition, which contains a total of 814,255 images evenly distributed across 62 classes.  The model architecture used for EMNIST follows the same CNN structure employed for MNIST.
\vspace{3pt}
\item \textbf{CIFAR-10} is a widely-used image classification dataset, containing a total of 60,000 color images categorized into 10 distinct classes. Each class consists of 6,000 images, providing a balanced distribution across the dataset. The images are relatively small, with dimensions of 32$\times$32 pixels and contain three RGB color channels.  In our experiments, we employ a robust CNN architecture, consisting of 6 convolutional layers followed by 3 fully connected layers.
\end{itemize}

\begin{figure}[t]

    \centering
    \begin{subfigure}[t]{0.1\textwidth}
        \centering
        \includegraphics[width=\textwidth,]{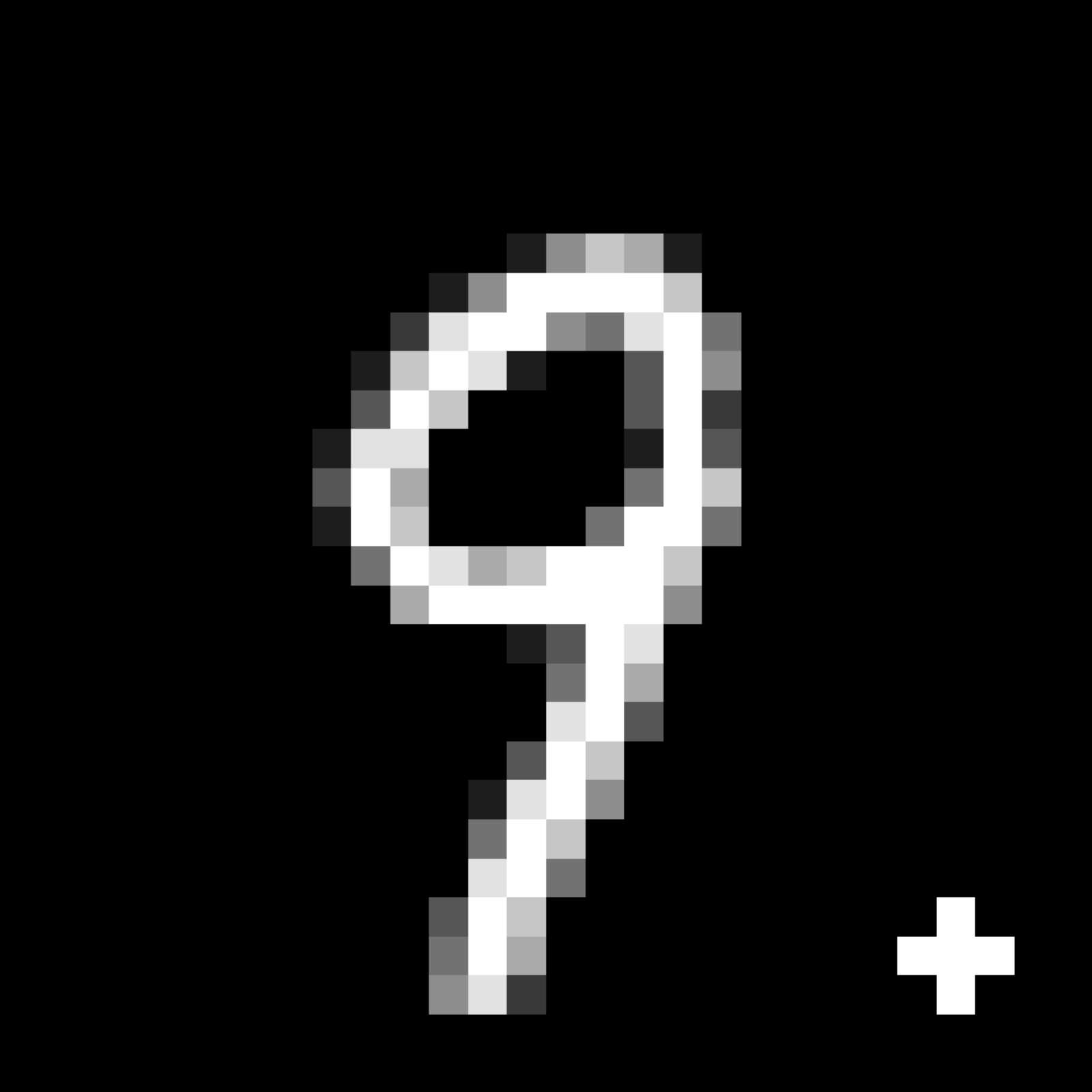}
        \caption{}
        \label{fig2:CBA}
    \end{subfigure}
    \hspace{40pt}
    \begin{subfigure}[t]{0.1\textwidth}
        \centering
        \includegraphics[width=\textwidth]{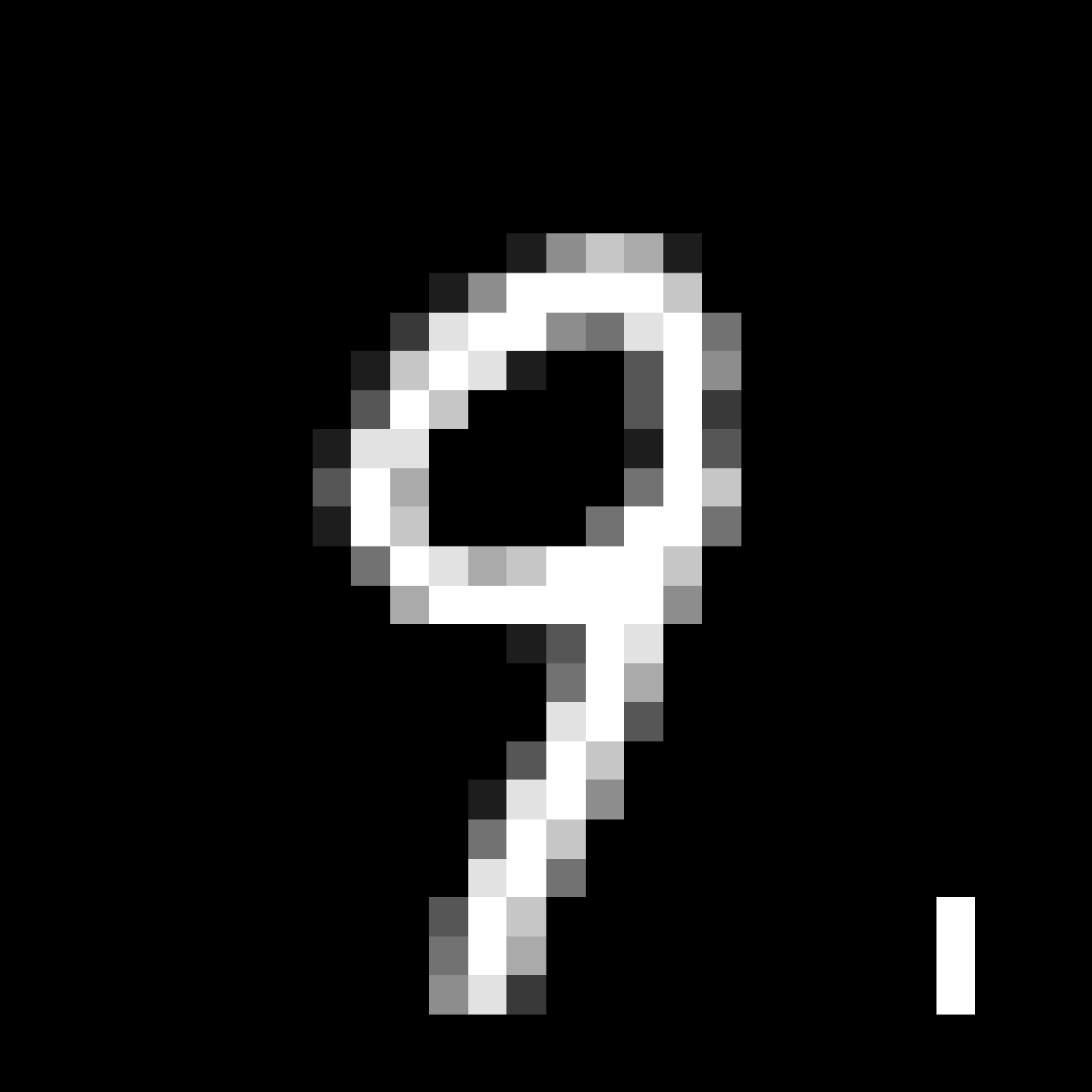}
        \caption{}
        \label{fig2:DBA1}
    \end{subfigure}
    \hspace{8pt}
    \begin{subfigure}[t]{0.1\textwidth}
        \centering
        \includegraphics[width=\textwidth]{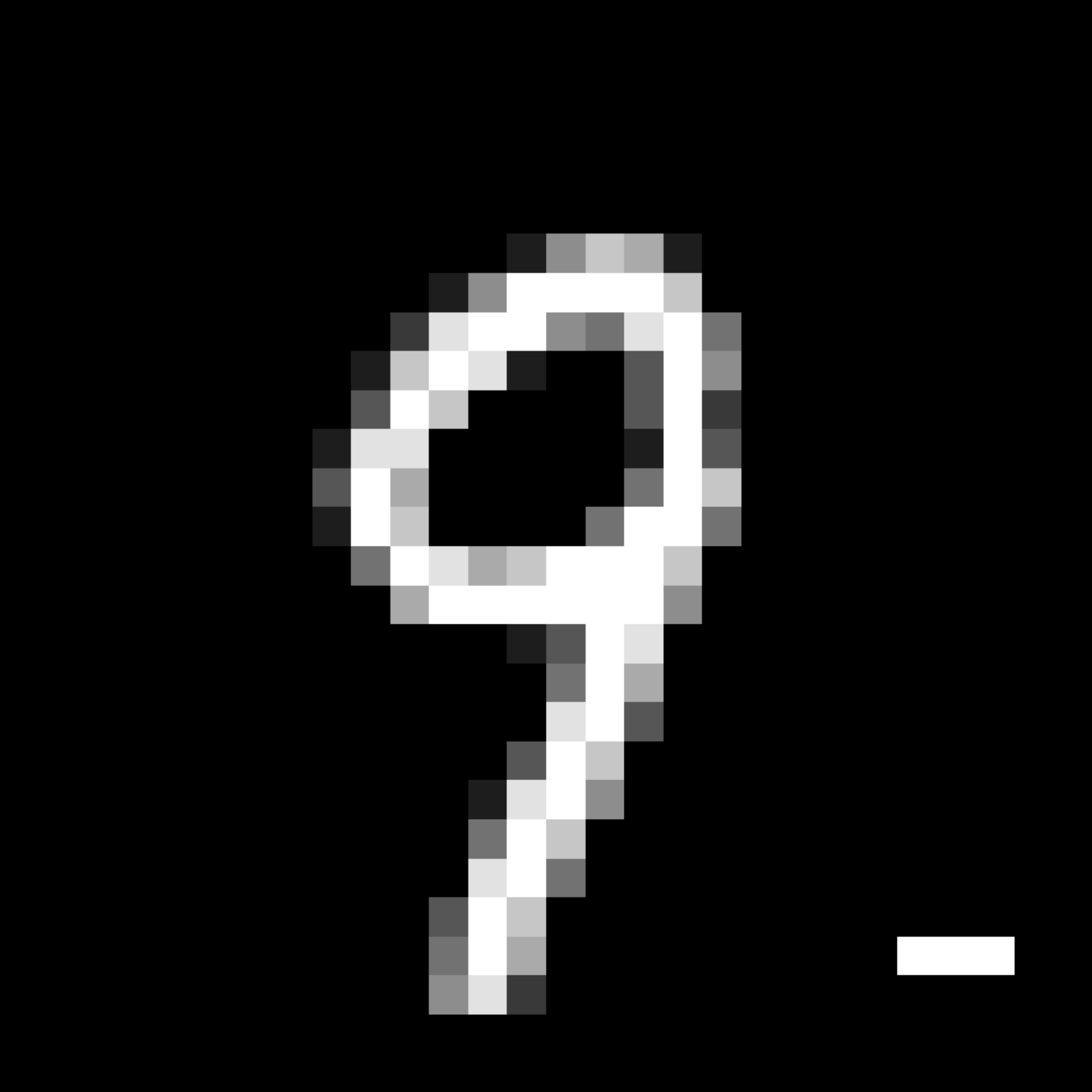}
        \caption{}
        \label{fig2:DBA2}
    \end{subfigure}
    \vspace{-5pt}
    \caption{CBA attack (a) and DBA attack (b, c).}
    \label{fig2:CBA and DBA}
\end{figure}

\noindent\textbf{Attack Settings.} We employ two distinct trigger patterns for injecting backdoor attacks: the CBA attack \cite{c:11} and the DBA attack \cite{c:36}. Figure \textbf{\ref{fig2:CBA and DBA}} provides a visual comparison of both patterns, specifically applied to class `9' from the MNIST dataset.

\begin{itemize} \item \textbf{Centralized Backdoor Attack (CBA):} In this attack, each malicious client embeds the entire trigger pattern into the data, meaning they have full access to and apply the complete backdoor trigger.
\vspace{3pt}
\item \textbf{Distributed Backdoor Attack (DBA):} In contrast, the DBA attack splits the trigger into smaller parts, with each malicious client embedding only a portion of the full trigger. This distributed approach increases the stealth of the attack by making detection more difficult. 
\end{itemize}

\begin{figure*}[tb]
    \centering
    \includegraphics[width=0.42\textwidth,height=0.28\textheight]{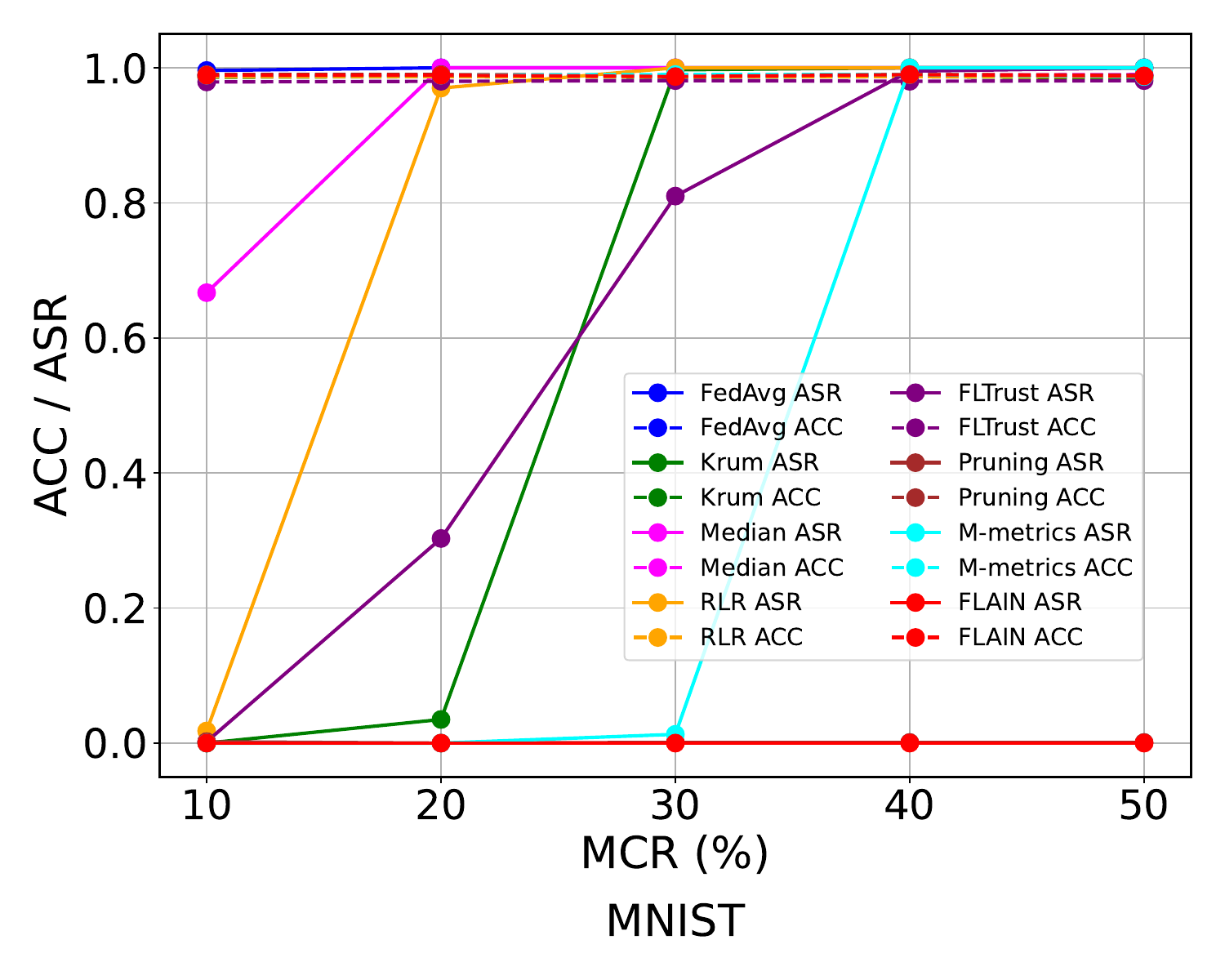}
    \hspace{50pt}
    \includegraphics[width=0.41\textwidth,height=0.28\textheight]{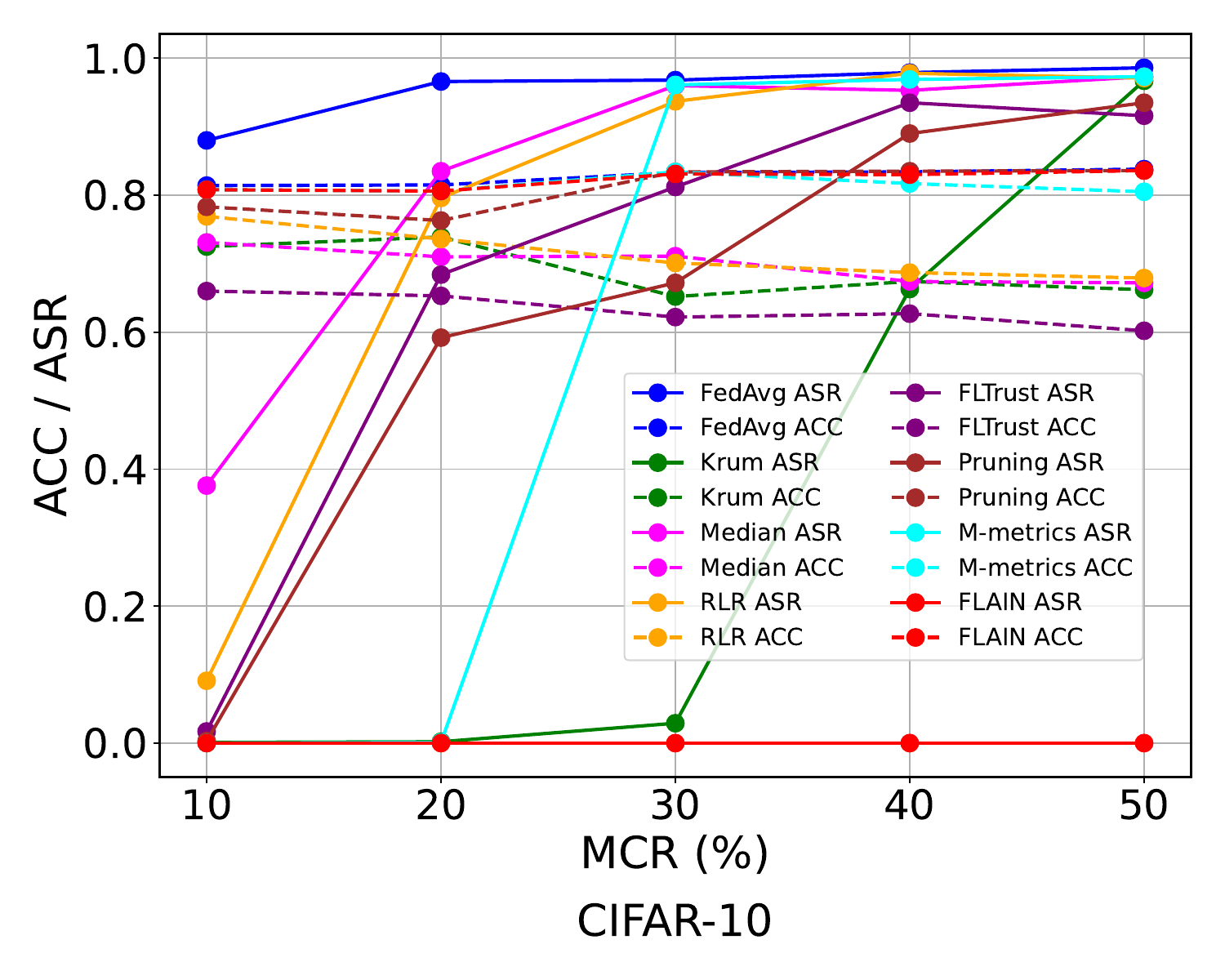}
    \vspace{-5pt}
    \caption{Performance under different MCR on MNIST and CIFAR-10.}
    \label{fig3:MCR}
\end{figure*}

\subsection{Experimental Results}
We set $step$ as 0.0001 and $\rho$ no more than 0.035 for all settings. The auxiliary dataset $\psi$ is sampled from the validation dataset, consisting of 60 images per class from CIFAR-10 and 20 images per class from MNIST, EMNIST, and FashionMNIST. We apply FLAIN to the first fully connected layer following the convolutional layers and conduct comparative experiments with several well-established defense techniques, including Krum \cite{c:13}, Median \cite{c:15}, RLR \cite{c:28}, FLTrust \cite{c:47}, M-metrics \cite{c:48}, and Pruning. In order to maintain consistency across the experimental setup, we also apply the Pruning method to the same first fully connected layer after the convolutional layers, ensuring a fair comparison between both FLAIN and Pruning methods.

\vspace{3pt}
\noindent\textbf{Defense  Performance under IID data.}   We configure 10 clients with the MCR set to 40\% and the poison data ratio (PDR) set to 30\%. We define different attack tasks using the format$(Ground -truth\,\, label,
Target \,\,label)$. For this evaluation, the CBA attack tasks are set to (0, 5), (0, 6), and (0, 7). 

As shown in Table \textbf{\ref{tab:table1}}, among the Byzantine-robust aggregation defense methods, Krum provides only marginal defense on the MNIST dataset, with the ASR remaining around 52\% across different attack configurations. Median fails to offer any meaningful defense under our attack settings. The RLR algorithm demonstrates promising defense performance on the MNIST, FashionMNIST, and EMNIST datasets, significantly reducing the ASR while maintaining model performance on clean data—particularly achieving an ASR of 0 on the MNIST dataset. However, on CIFAR-10, the complexity of the dataset and high MCR settings negatively affect RLR's defense capabilities, resulting in near-total failure. Similarly, FLTrust shows only modest defense on FashionMNIST and EMNIST, with average ASRs around 47.8\% and 35\%, respectively. M-metrics achieves robust defense performance only on the MNIST attack task, showing effective defense in the EMNIST attack tasks (0, 5) and (0, 7), but performs poorly on FashionMNIST and CIFAR-10. Both Pruning and FLAIN outperform other methods in terms of defense performance across most datasets, with Pruning delivering the best OPS in the MNIST (0, 7) attack setting. However, Pruning performs poorly on CIFAR-10, while FLAIN exhibits superior defense performance on this dataset. Specifically, FLAIN maintains an average ASR of just 0.13\%, with a minimal 0.6\% drop in ACC, while achieving the best OPS across all attack configurations.

\begin{table}[]
\caption{Performance of FLAIN under varying PDR.}
\vspace{-5pt}
\setlength{\tabcolsep}{0.7mm}
\begin{tabular}{|c|c|cc|cc|cc|}
\hline
\multirow{2}{*}{\textbf{Dataset}}  & \multirow{2}{*}{\textbf{Defense}} & \multicolumn{2}{c|}{20\%}        & \multicolumn{2}{c|}{40\%}        & \multicolumn{2}{c|}{60\%}        \\ \cline{3-8} 
                          &                          & \multicolumn{1}{c|}{ASR \textdownarrow} & ACC \textuparrow   & \multicolumn{1}{c|}{ASR \textdownarrow} & ACC \textuparrow   & \multicolumn{1}{c|}{ASR \textdownarrow} & ACC \textuparrow  \\ \hline
\multirow{2}{*}{FMNIST}   & FedAvg                   & 0.976                    & 0.908 & 0.982                    & 0.907 & 0.986                    & 0.910 \\ \cline{2-8} 
                          & FLAIN                    & 0.011                    & 0.883 & 0.006                    & 0.896 & 0.007                    & 0.892 \\ \hline
\multirow{2}{*}{CIFAR-10} & FedAvg                   & 0.926                    & 0.821 & 0.958                    & 0.812 & 0.970                     & 0.828 \\ \cline{2-8} 
                          & FLAIN                    & 0.031                    & 0.771 & 0.011                    & 0.793 & 0.010                     & 0.819 \\ \hline
\end{tabular}
\label{tab:table3}
\end{table}

\vspace{3pt}
\noindent\textbf{Defense Performance under Non-IID data.}
 We utilize the Dirichlet distribution \cite{c:49} with a concentration parameter $\alpha = 0.5$ to simulate Non-IID data distribution scenarios, which are often encountered in real-world FL environments. In our evaluation, the DBA attack is applied across a range of datasets to assess the robustness of defense methods under Non-IID conditions. All other experimental settings remain consistent with IID conditions.

As shown in Table \textbf{\ref{tab:table2}}, the Krum algorithm achieves effective defense results only on the EMNIST dataset, where the ASR is controlled at 0 under different attack settings, but the average ACC decreases by 12.9\%. In contrast, the Median algorithm fails to provide effective defense across multiple datasets under various attack settings. The RLR algorithm performs well on the MNIST, FashionMNIST, and CIFAR-10 datasets, although the ACC loss is more significant on FashionMNIST. On CIFAR-10, RLR nearly fails due to considerable discrepancies in data distribution between clients in the Non-IID setting, which result in substantial variations in locally uploaded updates, thereby increasing the error when the server performs directional voting on each update dimension. FLTrust demonstrates some defense effects under the (0, 5) and (0, 6) attack settings on the EMNIST dataset, but still sacrifices a significant portion of the model's benign performance. M-metrics achieves strong defense performance on the EMNIST dataset, with the overall ASR controlled at an average of 2.3\%, and the average ACC only decreasing by 2.4\%. The Pruning algorithm is less affected by data distribution on the MNIST and EMNIST datasets, particularly on EMNIST, where it achieves favorable OPS. FLAIN outperforms other methods in defense effectiveness across various datasets, especially on CIFAR-10. While other defense methods nearly fail, FLAIN limits the maximum ACC loss to within 5\% and reduces the ASR to approximately 1.8\%.

\begin{figure}[tb]
    \centering
    \includegraphics[width=0.4\textwidth,height=0.26\textheight]{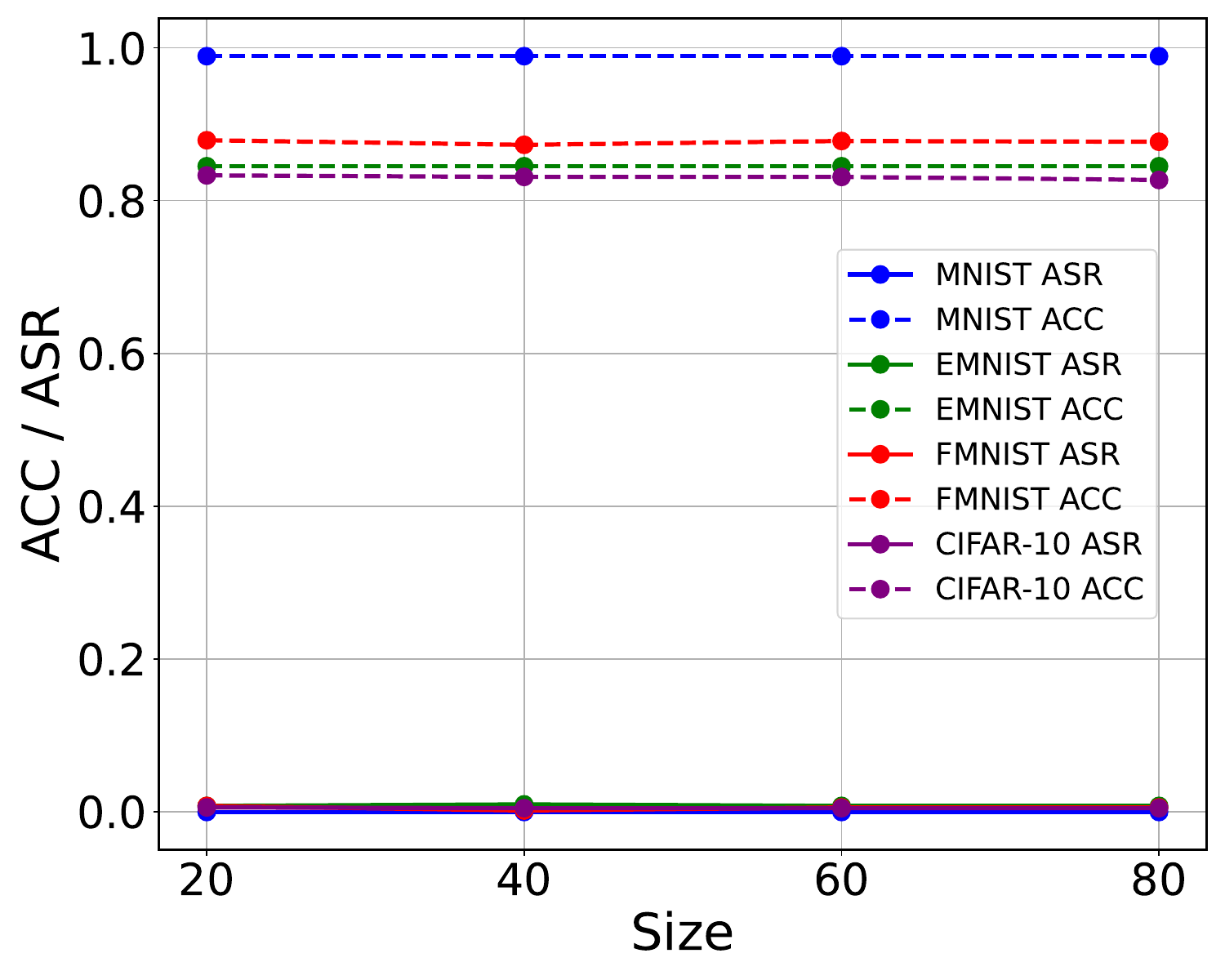}
    \vspace{-5pt}
\caption{Impact of auxiliary data sizes on performance.}
\label{fig4:Size}
\end{figure}

\vspace{3pt}
\noindent\textbf{Performance under Different MCR.} We configure 20 clients, set the PDR to 50\%, and apply a CBA attack task set to (0, 5). To evaluate the impact of different MCR values, we conduct experiments with MCR settings of 10\%, 20\%, 30\%, 40\%, and 50\%.

As depicted in Figure \textbf{\ref{fig3:MCR}}, the defense performance of various methods is evaluated across different MCR settings for both MNIST and CIFAR-10. At an MCR of 10\%, Krum, RLR, FLTrust, Pruning, and FLAIN all demonstrate effective defense. However, as the MCR increases, the performance of Krum, RLR, and FLTrust declines significantly. M-metrics remains effective below a 30\% MCR, successfully limiting the ASR and demonstrating consistent robustness. However, once the MCR exceeds 30\%, the ASR begins to rise, reflecting a gradual decline in defense effectiveness. In contrast, Pruning and FLAIN maintain stable defense capabilities with minimal impact from the increasing MCR. On CIFAR-10, Krum, RLR, FLTrust, and Pruning successfully reduce the ASR at an MCR of 10\%, but they experience substantial ACC losses. M-metrics performs well when the MCR is below 20\%, maintaining a low ASR and effectively defending against attacks. As the MCR surpasses 20\%, the ASR increases, indicating a decrease in defense effectiveness. As MCR rises, the defense performance of these methods sharply declines, with Pruning showing weaker robustness compared to its performance on MNIST. In contrast, FLAIN remains relatively unaffected by changes in MCR, keeping the ASR near 0 and experiencing only a slight reduction in ACC.

\vspace{4pt}
\noindent\textbf{Impact of Varying PDR.} We configure 20 clients with the MCR set to 30\%, define the DBA attack task as (0, 6), and adjust the PDR to 20\%, 40\%, and 60\%. Using both the FashionMNIST and CIFAR-10 datasets, we evaluate the impact of varying PDR settings on the defense performance of the FLAIN.

As presented in Table \textbf{\ref{tab:table3}}, on the MNIST, as the PDR increases, the ASR remains consistently controlled at an average of 0.8\%, while the model's performance on benign tasks shows a slight average decrease in ACC by 1.8\%. Similarly, on the CIFAR-10, FLAIN exhibits slightly weaker defense effectiveness when the PDR is set at 20\%. However, as the PDR is raised to 40\% and 60\%, the defense becomes more effective, with the ASR reduced to approximately 1\%, while the ACC reduction is kept within 2\%. Considering both the FashionMNIST and CIFAR-10 datasets together, FLAIN demonstrates minimal sensitivity to variations in PDR.

\vspace{4pt}
\noindent\textbf{Impact of Varying Auxiliary Data Sizes.} We configure 20 clients with an MCR of 30\%, define the CBA attack task as (0, 7), and set the PDR to 30\%. The sample sizes for each class are set to 20, 40, 60, and 80 across different datasets to evaluate the impact of varying auxiliary dataset sizes on the defense performance of the FLAIN.

As shown in Figure \textbf{\ref{fig4:Size}}, FLAIN demonstrates ideal defense performance even with only 20 samples per class. As the auxiliary sample size increases, the improvements in defense performance (ASR) and benign performance (ACC) become marginal, suggesting that FLAIN can achieve satisfactory defense results with a relatively small number of useful auxiliary samples.

\section{Conclusion and Future Work}
In this paper, we introduce FLAIN, a backdoor defense method that mitigates attacks by flipping the weight updates of neurons with low-activation inputs. We propose a performance-adaptive threshold that automatically balances the tradeoff between eliminating backdoors and maintaining accuracy on clean data. The synergy between flipping and the performance-adaptive threshold enhances FLAIN's effectiveness. In evaluations, FLAIN demonstrates superior adaptability compared to direct pruning methods across a wide range of attack scenarios, including Non-IID data distributions and high MCR settings. It consistently reduces the impact of backdoors while preserving the model's performance on the main task. Future work will explore further improvements to FLAIN, including its application in data-free scenarios and with more complex models.

\section*{Acknowledgments}

This work was supported by the National Natural Science Foundation of China (U2441285, 62222605) and the Natural Science Foundation of Jiangsu Province of China (BK20222012).
\balance
\bibliographystyle{ACM-Reference-Format}
\bibliography{sample-base}


\begin{thebibliography}{49}


\ifx \showCODEN    \undefined \def \showCODEN     #1{\unskip}     \fi
\ifx \showISBNx    \undefined \def \showISBNx     #1{\unskip}     \fi
\ifx \showISBNxiii \undefined \def \showISBNxiii  #1{\unskip}     \fi
\ifx \showISSN     \undefined \def \showISSN      #1{\unskip}     \fi
\ifx \showLCCN     \undefined \def \showLCCN      #1{\unskip}     \fi
\ifx \shownote     \undefined \def \shownote      #1{#1}          \fi
\ifx \showarticletitle \undefined \def \showarticletitle #1{#1}   \fi
\ifx \showURL      \undefined \def \showURL       {\relax}        \fi
\providecommand\bibfield[2]{#2}
\providecommand\bibinfo[2]{#2}
\providecommand\natexlab[1]{#1}
\providecommand\showeprint[2][]{arXiv:#2}

\bibitem[Alzubi et~al\mbox{.}(2022)]%
        {c:2}
\bibfield{author}{\bibinfo{person}{Jafar~A Alzubi}, \bibinfo{person}{Omar~A Alzubi}, \bibinfo{person}{Ashish Singh}, {and} \bibinfo{person}{Manikandan Ramachandran}.} \bibinfo{year}{2022}\natexlab{}.
\newblock \showarticletitle{Cloud-IIoT-based electronic health record privacy-preserving by CNN and blockchain-enabled federated learning}.
\newblock \bibinfo{journal}{\emph{IEEE Transactions on Industrial Informatics}} \bibinfo{volume}{19}, \bibinfo{number}{1} (\bibinfo{year}{2022}), \bibinfo{pages}{1080--1087}.
\newblock


\bibitem[Aramoon et~al\mbox{.}(2021)]%
        {c:31}
\bibfield{author}{\bibinfo{person}{Omid Aramoon}, \bibinfo{person}{Pin-Yu Chen}, \bibinfo{person}{Gang Qu}, {and} \bibinfo{person}{Yuan Tian}.} \bibinfo{year}{2021}\natexlab{}.
\newblock \showarticletitle{Meta Federated Learning}.
\newblock \bibinfo{journal}{\emph{arXiv preprint arXiv:2102.05561}} (\bibinfo{year}{2021}).
\newblock


\bibitem[Bagdasaryan et~al\mbox{.}(2020)]%
        {c:11}
\bibfield{author}{\bibinfo{person}{Eugene Bagdasaryan}, \bibinfo{person}{Andreas Veit}, \bibinfo{person}{Yiqing Hua}, \bibinfo{person}{Deborah Estrin}, {and} \bibinfo{person}{Vitaly Shmatikov}.} \bibinfo{year}{2020}\natexlab{}.
\newblock \showarticletitle{How to backdoor federated learning}. In \bibinfo{booktitle}{\emph{Proceedings of the International Conference on Artificial Intelligence and Statistics}}. \bibinfo{pages}{2938--2948}.
\newblock


\bibitem[Basu et~al\mbox{.}(2021)]%
        {c:4}
\bibfield{author}{\bibinfo{person}{Priyam Basu}, \bibinfo{person}{Tiasa~Singha Roy}, \bibinfo{person}{Rakshit Naidu}, {and} \bibinfo{person}{Zumrut Muftuoglu}.} \bibinfo{year}{2021}\natexlab{}.
\newblock \showarticletitle{Privacy enabled financial text classification using differential privacy and federated learning}.
\newblock \bibinfo{journal}{\emph{arXiv preprint arXiv:2110.01643}} (\bibinfo{year}{2021}).
\newblock


\bibitem[Bhagoji et~al\mbox{.}(2019)]%
        {c:21}
\bibfield{author}{\bibinfo{person}{Arjun~Nitin Bhagoji}, \bibinfo{person}{Supriyo Chakraborty}, \bibinfo{person}{Prateek Mittal}, {and} \bibinfo{person}{Seraphin Calo}.} \bibinfo{year}{2019}\natexlab{}.
\newblock \showarticletitle{Analyzing federated learning through an adversarial lens}. In \bibinfo{booktitle}{\emph{Proceedings of the International Conference on Machine Learning}}. \bibinfo{pages}{634--643}.
\newblock


\bibitem[Blanchard et~al\mbox{.}(2017)]%
        {c:13}
\bibfield{author}{\bibinfo{person}{Peva Blanchard}, \bibinfo{person}{El~Mahdi El~Mhamdi}, \bibinfo{person}{Rachid Guerraoui}, {and} \bibinfo{person}{Julien Stainer}.} \bibinfo{year}{2017}\natexlab{}.
\newblock \showarticletitle{Machine learning with adversaries: Byzantine tolerant gradient descent}. In \bibinfo{booktitle}{\emph{Advances in neural information processing systems}}, Vol.~\bibinfo{volume}{30}.
\newblock


\bibitem[Cao et~al\mbox{.}(2020a)]%
        {c:37}
\bibfield{author}{\bibinfo{person}{Xiaoyu Cao}, \bibinfo{person}{Minghong Fang}, \bibinfo{person}{Jia Liu}, {and} \bibinfo{person}{Neil~Zhenqiang Gong}.} \bibinfo{year}{2020}\natexlab{a}.
\newblock \showarticletitle{Fltrust: Byzantine-robust federated learning via trust bootstrapping}.
\newblock \bibinfo{journal}{\emph{arXiv preprint arXiv:2012.13995}} (\bibinfo{year}{2020}).
\newblock


\bibitem[Cao et~al\mbox{.}(2020b)]%
        {c:47}
\bibfield{author}{\bibinfo{person}{Xiaoyu Cao}, \bibinfo{person}{Minghong Fang}, \bibinfo{person}{Jia Liu}, {and} \bibinfo{person}{Neil~Zhenqiang Gong}.} \bibinfo{year}{2020}\natexlab{b}.
\newblock \showarticletitle{Fltrust: Byzantine-robust federated learning via trust bootstrapping}.
\newblock \bibinfo{journal}{\emph{arXiv preprint arXiv:2012.13995}} (\bibinfo{year}{2020}).
\newblock


\bibitem[Chatterjee et~al\mbox{.}(2023)]%
        {c:5}
\bibfield{author}{\bibinfo{person}{Pushpita Chatterjee}, \bibinfo{person}{Debashis Das}, {and} \bibinfo{person}{Danda~B Rawat}.} \bibinfo{year}{2023}\natexlab{}.
\newblock \showarticletitle{Next generation financial services: Role of blockchain enabled federated learning and metaverse}. In \bibinfo{booktitle}{\emph{Proceedings of the IEEE/ACM 23rd International Symposium on Cluster, Cloud and Internet Computing Workshops}}. \bibinfo{pages}{69--74}.
\newblock


\bibitem[Chen et~al\mbox{.}(2020)]%
        {c:40}
\bibfield{author}{\bibinfo{person}{Zheyi Chen}, \bibinfo{person}{Pu Tian}, \bibinfo{person}{Weixian Liao}, {and} \bibinfo{person}{Wei Yu}.} \bibinfo{year}{2020}\natexlab{}.
\newblock \showarticletitle{Zero knowledge clustering based adversarial mitigation in heterogeneous federated learning}.
\newblock \bibinfo{journal}{\emph{IEEE Transactions on Network Science and Engineering}} \bibinfo{volume}{8}, \bibinfo{number}{2} (\bibinfo{year}{2020}), \bibinfo{pages}{1070--1083}.
\newblock


\bibitem[Cohen et~al\mbox{.}(2017)]%
        {c:45}
\bibfield{author}{\bibinfo{person}{Gregory Cohen}, \bibinfo{person}{Saeed Afshar}, \bibinfo{person}{Jonathan Tapson}, {and} \bibinfo{person}{Andre Van~Schaik}.} \bibinfo{year}{2017}\natexlab{}.
\newblock \showarticletitle{EMNIST: Extending MNIST to handwritten letters}. In \bibinfo{booktitle}{\emph{Proceedings of the International Joint Conference on Neural Networks}}. \bibinfo{pages}{2921--2926}.
\newblock


\bibitem[Deng(2012)]%
        {c:43}
\bibfield{author}{\bibinfo{person}{Li Deng}.} \bibinfo{year}{2012}\natexlab{}.
\newblock \showarticletitle{The mnist database of handwritten digit images for machine learning research}.
\newblock \bibinfo{journal}{\emph{IEEE Signal Processing Magazine}} \bibinfo{volume}{29}, \bibinfo{number}{6} (\bibinfo{year}{2012}), \bibinfo{pages}{141--142}.
\newblock


\bibitem[Fang and Chen(2023)]%
        {c:9}
\bibfield{author}{\bibinfo{person}{Pei Fang} {and} \bibinfo{person}{Jinghui Chen}.} \bibinfo{year}{2023}\natexlab{}.
\newblock \showarticletitle{On the vulnerability of backdoor defenses for federated learning}. In \bibinfo{booktitle}{\emph{Proceedings of the AAAI Conference on Artificial Intelligence}}, Vol.~\bibinfo{volume}{37}. \bibinfo{pages}{11800--11808}.
\newblock


\bibitem[Fung et~al\mbox{.}(2018)]%
        {c:19}
\bibfield{author}{\bibinfo{person}{Clement Fung}, \bibinfo{person}{Chris~JM Yoon}, {and} \bibinfo{person}{Ivan Beschastnikh}.} \bibinfo{year}{2018}\natexlab{}.
\newblock \showarticletitle{Mitigating sybils in federated learning poisoning}.
\newblock \bibinfo{journal}{\emph{arXiv preprint arXiv:1808.04866}} (\bibinfo{year}{2018}).
\newblock


\bibitem[Gu et~al\mbox{.}(2017)]%
        {c:22}
\bibfield{author}{\bibinfo{person}{Tianyu Gu}, \bibinfo{person}{Brendan Dolan-Gavitt}, {and} \bibinfo{person}{Siddharth Garg}.} \bibinfo{year}{2017}\natexlab{}.
\newblock \showarticletitle{Badnets: Identifying vulnerabilities in the machine learning model supply chain}.
\newblock \bibinfo{journal}{\emph{arXiv preprint arXiv:1708.06733}} (\bibinfo{year}{2017}).
\newblock


\bibitem[Huang et~al\mbox{.}(2023a)]%
        {c:30}
\bibfield{author}{\bibinfo{person}{Siquan Huang}, \bibinfo{person}{Yijiang Li}, \bibinfo{person}{Chong Chen}, \bibinfo{person}{Leyu Shi}, {and} \bibinfo{person}{Ying Gao}.} \bibinfo{year}{2023}\natexlab{a}.
\newblock \showarticletitle{Multi-metrics adaptively identifies backdoors in Federated learning}. In \bibinfo{booktitle}{\emph{Proceedings of the IEEE/CVF International Conference on Computer Vision}}. \bibinfo{pages}{4652--4662}.
\newblock


\bibitem[Huang et~al\mbox{.}(2023b)]%
        {c:48}
\bibfield{author}{\bibinfo{person}{Siquan Huang}, \bibinfo{person}{Yijiang Li}, \bibinfo{person}{Chong Chen}, \bibinfo{person}{Leyu Shi}, {and} \bibinfo{person}{Ying Gao}.} \bibinfo{year}{2023}\natexlab{b}.
\newblock \showarticletitle{Multi-metrics adaptively identifies backdoors in Federated learning}. In \bibinfo{booktitle}{\emph{Proceedings of the IEEE/CVF International Conference on Computer Vision}}. \bibinfo{pages}{4652--4662}.
\newblock


\bibitem[Kingma and Ba(2014)]%
        {c:42}
\bibfield{author}{\bibinfo{person}{Diederik~P Kingma} {and} \bibinfo{person}{Jimmy Ba}.} \bibinfo{year}{2014}\natexlab{}.
\newblock \showarticletitle{Adam: A method for stochastic optimization}.
\newblock \bibinfo{journal}{\emph{arXiv preprint arXiv:1412.6980}} (\bibinfo{year}{2014}).
\newblock


\bibitem[Krizhevsky et~al\mbox{.}(2009)]%
        {c:46}
\bibfield{author}{\bibinfo{person}{Alex Krizhevsky}, \bibinfo{person}{Geoffrey Hinton}, {et~al\mbox{.}}} \bibinfo{year}{2009}\natexlab{}.
\newblock \showarticletitle{Learning multiple layers of features from tiny images}.
\newblock  (\bibinfo{year}{2009}).
\newblock


\bibitem[Liang et~al\mbox{.}(2023)]%
        {c:33}
\bibfield{author}{\bibinfo{person}{Xianfeng Liang}, \bibinfo{person}{Shuheng Shen}, \bibinfo{person}{Enhong Chen}, \bibinfo{person}{Jinchang Liu}, \bibinfo{person}{Qi Liu}, \bibinfo{person}{Yifei Cheng}, {and} \bibinfo{person}{Zhen Pan}.} \bibinfo{year}{2023}\natexlab{}.
\newblock \showarticletitle{Accelerating local SGD for non-IID data using variance reduction}.
\newblock \bibinfo{journal}{\emph{Frontiers of Computer Science}} \bibinfo{volume}{17}, \bibinfo{number}{2} (\bibinfo{year}{2023}), \bibinfo{pages}{172311}.
\newblock


\bibitem[Lin(2016)]%
        {c:49}
\bibfield{author}{\bibinfo{person}{Jiayu Lin}.} \bibinfo{year}{2016}\natexlab{}.
\newblock \showarticletitle{On the dirichlet distribution}.
\newblock \bibinfo{journal}{\emph{Department of Mathematics and Statistics, Queens University}}  \bibinfo{volume}{40} (\bibinfo{year}{2016}).
\newblock


\bibitem[Liu et~al\mbox{.}(2018)]%
        {c:23}
\bibfield{author}{\bibinfo{person}{Kang Liu}, \bibinfo{person}{Brendan Dolan-Gavitt}, {and} \bibinfo{person}{Siddharth Garg}.} \bibinfo{year}{2018}\natexlab{}.
\newblock \showarticletitle{Fine-pruning: Defending against backdooring attacks on deep neural networks}. In \bibinfo{booktitle}{\emph{Proceedings of the International Symposium on Research in Attacks, Intrusions, and Defenses}}. \bibinfo{pages}{273--294}.
\newblock


\bibitem[Manias and Shami(2021)]%
        {c:6}
\bibfield{author}{\bibinfo{person}{Dimitrios~Michael Manias} {and} \bibinfo{person}{Abdallah Shami}.} \bibinfo{year}{2021}\natexlab{}.
\newblock \showarticletitle{Making a case for federated learning in the internet of vehicles and intelligent transportation systems}.
\newblock \bibinfo{journal}{\emph{IEEE Network}} \bibinfo{volume}{35}, \bibinfo{number}{3} (\bibinfo{year}{2021}), \bibinfo{pages}{88--94}.
\newblock


\bibitem[McMahan et~al\mbox{.}(2017)]%
        {c:1}
\bibfield{author}{\bibinfo{person}{Brendan McMahan}, \bibinfo{person}{Eider Moore}, \bibinfo{person}{Daniel Ramage}, \bibinfo{person}{Seth Hampson}, {and} \bibinfo{person}{Blaise~Aguera y Arcas}.} \bibinfo{year}{2017}\natexlab{}.
\newblock \showarticletitle{{Communication-efficient learning of deep networks from decentralized data}}. In \bibinfo{booktitle}{\emph{Proceedings of the International Conference on Artificial Intelligence and Statistics}}, Vol.~\bibinfo{volume}{54}. \bibinfo{pages}{1273–1282}.
\newblock


\bibitem[Mhamdi et~al\mbox{.}(2018)]%
        {c:14}
\bibfield{author}{\bibinfo{person}{El~Mahdi~El Mhamdi}, \bibinfo{person}{Rachid Guerraoui}, {and} \bibinfo{person}{Sébastien Rouault}.} \bibinfo{year}{2018}\natexlab{}.
\newblock \showarticletitle{The Hidden Vulnerability of Distributed Learning in Byzantium}.
\newblock \bibinfo{journal}{\emph{arXiv preprint arXiv:1802.07927}} (\bibinfo{year}{2018}).
\newblock


\bibitem[Nguyen et~al\mbox{.}(2021)]%
        {c:29}
\bibfield{author}{\bibinfo{person}{Thien~Duc Nguyen}, \bibinfo{person}{Phillip Rieger}, \bibinfo{person}{Hossein Yalame}, \bibinfo{person}{Helen M{\"o}llering}, \bibinfo{person}{Hossein Fereidooni}, \bibinfo{person}{Samuel Marchal}, \bibinfo{person}{Markus Miettinen}, \bibinfo{person}{Azalia Mirhoseini}, \bibinfo{person}{Ahmad-Reza Sadeghi}, \bibinfo{person}{T. Schneider}, {and} \bibinfo{person}{Shaza Zeitouni}.} \bibinfo{year}{2021}\natexlab{}.
\newblock \showarticletitle{FLGUARD: Secure and Private Federated Learning}.
\newblock \bibinfo{journal}{\emph{arXiv preprint arXiv:2101.02281}} (\bibinfo{year}{2021}).
\newblock


\bibitem[Ozdayi et~al\mbox{.}(2021)]%
        {c:28}
\bibfield{author}{\bibinfo{person}{Mustafa~Safa Ozdayi}, \bibinfo{person}{Murat Kantarcioglu}, {and} \bibinfo{person}{Yulia~R Gel}.} \bibinfo{year}{2021}\natexlab{}.
\newblock \showarticletitle{Defending against backdoors in federated learning with robust learning rate}. In \bibinfo{booktitle}{\emph{Proceedings of the AAAI Conference on Artificial Intelligence}}. \bibinfo{pages}{9268--9276}.
\newblock


\bibitem[Qin et~al\mbox{.}(2024)]%
        {c:18}
\bibfield{author}{\bibinfo{person}{Zhen Qin}, \bibinfo{person}{Feiyi Chen}, \bibinfo{person}{Chen Zhi}, \bibinfo{person}{Xueqiang Yan}, {and} \bibinfo{person}{Shuiguang Deng}.} \bibinfo{year}{2024}\natexlab{}.
\newblock \showarticletitle{Resisting Backdoor Attacks in Federated Learning via Bidirectional Elections and Individual Perspective}. In \bibinfo{booktitle}{\emph{Proceedings of the AAAI Conference on Artificial Intelligence}}. \bibinfo{pages}{14677--14685}.
\newblock


\bibitem[Ramachandran et~al\mbox{.}(2017)]%
        {c:41}
\bibfield{author}{\bibinfo{person}{Prajit Ramachandran}, \bibinfo{person}{Barret Zoph}, {and} \bibinfo{person}{Quoc~V Le}.} \bibinfo{year}{2017}\natexlab{}.
\newblock \showarticletitle{Searching for activation functions}.
\newblock \bibinfo{journal}{\emph{arXiv preprint arXiv:1710.05941}} (\bibinfo{year}{2017}).
\newblock


\bibitem[Salim and Park(2022)]%
        {c:3}
\bibfield{author}{\bibinfo{person}{Mikail~Mohammed Salim} {and} \bibinfo{person}{Jong~Hyuk Park}.} \bibinfo{year}{2022}\natexlab{}.
\newblock \showarticletitle{Federated learning-based secure electronic health record sharing scheme in medical informatics}.
\newblock \bibinfo{journal}{\emph{IEEE Journal of Biomedical and Health Informatics}} \bibinfo{volume}{27}, \bibinfo{number}{2} (\bibinfo{year}{2022}), \bibinfo{pages}{617--624}.
\newblock


\bibitem[Sun et~al\mbox{.}(2019)]%
        {c:10}
\bibfield{author}{\bibinfo{person}{Ziteng Sun}, \bibinfo{person}{Peter Kairouz}, \bibinfo{person}{Ananda~Theertha Suresh}, {and} \bibinfo{person}{H~Brendan McMahan}.} \bibinfo{year}{2019}\natexlab{}.
\newblock \showarticletitle{Can you really backdoor federated learning?}
\newblock \bibinfo{journal}{\emph{arXiv preprint arXiv:1911.07963}} (\bibinfo{year}{2019}).
\newblock


\bibitem[Tan et~al\mbox{.}(2020)]%
        {c:39}
\bibfield{author}{\bibinfo{person}{Junjie Tan}, \bibinfo{person}{Ying-Chang Liang}, \bibinfo{person}{Nguyen~Cong Luong}, {and} \bibinfo{person}{Dusit Niyato}.} \bibinfo{year}{2020}\natexlab{}.
\newblock \showarticletitle{Toward smart security enhancement of federated learning networks}.
\newblock \bibinfo{journal}{\emph{IEEE Network}} \bibinfo{volume}{35}, \bibinfo{number}{1} (\bibinfo{year}{2020}), \bibinfo{pages}{340--347}.
\newblock


\bibitem[Wang et~al\mbox{.}(2019)]%
        {c:24}
\bibfield{author}{\bibinfo{person}{Bolun Wang}, \bibinfo{person}{Yuanshun Yao}, \bibinfo{person}{Shawn Shan}, \bibinfo{person}{Huiying Li}, \bibinfo{person}{Bimal Viswanath}, \bibinfo{person}{Haitao Zheng}, {and} \bibinfo{person}{Ben~Y Zhao}.} \bibinfo{year}{2019}\natexlab{}.
\newblock \showarticletitle{Neural cleanse: Identifying and mitigating backdoor attacks in neural networks}. In \bibinfo{booktitle}{\emph{Proceedings of the IEEE Symposium on Security and Privacy}}. \bibinfo{pages}{707--723}.
\newblock


\bibitem[Wang et~al\mbox{.}(2020a)]%
        {c:34}
\bibfield{author}{\bibinfo{person}{Hongyi Wang}, \bibinfo{person}{Kartik Sreenivasan}, \bibinfo{person}{Shashank Rajput}, \bibinfo{person}{Harit Vishwakarma}, \bibinfo{person}{Saurabh Agarwal}, \bibinfo{person}{Jy-yong Sohn}, \bibinfo{person}{Kangwook Lee}, {and} \bibinfo{person}{Dimitris Papailiopoulos}.} \bibinfo{year}{2020}\natexlab{a}.
\newblock \showarticletitle{Attack of the tails: Yes, you really can backdoor federated learning}. In \bibinfo{booktitle}{\emph{Advances in Neural Information Processing Systems}}, Vol.~\bibinfo{volume}{33}. \bibinfo{pages}{16070--16084}.
\newblock


\bibitem[Wang et~al\mbox{.}(2020b)]%
        {c:38}
\bibfield{author}{\bibinfo{person}{Yuao Wang}, \bibinfo{person}{Tianqing Zhu}, \bibinfo{person}{Wenhan Chang}, \bibinfo{person}{Sheng Shen}, {and} \bibinfo{person}{Wei Ren}.} \bibinfo{year}{2020}\natexlab{b}.
\newblock \showarticletitle{Model poisoning defense on federated learning: A validation based approach}. In \bibinfo{booktitle}{\emph{Proceedings of the International Conference on Network and System Security}}. \bibinfo{pages}{207--223}.
\newblock


\bibitem[Wu et~al\mbox{.}(2020)]%
        {c:32}
\bibfield{author}{\bibinfo{person}{Chen Wu}, \bibinfo{person}{Xian Yang}, \bibinfo{person}{Sencun Zhu}, {and} \bibinfo{person}{Prasenjit Mitra}.} \bibinfo{year}{2020}\natexlab{}.
\newblock \showarticletitle{Mitigating backdoor attacks in federated learning}.
\newblock \bibinfo{journal}{\emph{arXiv preprint arXiv:2011.01767}} (\bibinfo{year}{2020}).
\newblock


\bibitem[Wu and Wang(2021)]%
        {c:27}
\bibfield{author}{\bibinfo{person}{Dongxian Wu} {and} \bibinfo{person}{Yisen Wang}.} \bibinfo{year}{2021}\natexlab{}.
\newblock \showarticletitle{Adversarial neuron pruning purifies backdoored deep models}. In \bibinfo{booktitle}{\emph{Advances in Neural Information Processing Systems}}, Vol.~\bibinfo{volume}{34}. \bibinfo{pages}{16913--16925}.
\newblock


\bibitem[Xiao et~al\mbox{.}(2017)]%
        {c:44}
\bibfield{author}{\bibinfo{person}{Han Xiao}, \bibinfo{person}{Kashif Rasul}, {and} \bibinfo{person}{Roland Vollgraf}.} \bibinfo{year}{2017}\natexlab{}.
\newblock \showarticletitle{Fashion-mnist: a novel image dataset for benchmarking machine learning algorithms}.
\newblock \bibinfo{journal}{\emph{arXiv preprint arXiv:1708.07747}} (\bibinfo{year}{2017}).
\newblock


\bibitem[Xie et~al\mbox{.}(2019)]%
        {c:36}
\bibfield{author}{\bibinfo{person}{Chulin Xie}, \bibinfo{person}{Keli Huang}, \bibinfo{person}{Pin-Yu Chen}, {and} \bibinfo{person}{Bo Li}.} \bibinfo{year}{2019}\natexlab{}.
\newblock \showarticletitle{Dba: Distributed backdoor attacks against federated learning}. In \bibinfo{booktitle}{\emph{Proceedings of the International Conference on Learning Representations}}.
\newblock


\bibitem[Xie et~al\mbox{.}(2020)]%
        {c:20}
\bibfield{author}{\bibinfo{person}{Cong Xie}, \bibinfo{person}{Oluwasanmi Koyejo}, {and} \bibinfo{person}{Indranil Gupta}.} \bibinfo{year}{2020}\natexlab{}.
\newblock \showarticletitle{Fall of empires: Breaking byzantine-tolerant sgd by inner product manipulation}. In \bibinfo{booktitle}{\emph{Proceedings of the 35th Uncertainty in Artificial Intelligence Conference}}. \bibinfo{pages}{261--270}.
\newblock


\bibitem[Yin et~al\mbox{.}(2018)]%
        {c:15}
\bibfield{author}{\bibinfo{person}{Dong Yin}, \bibinfo{person}{Yudong Chen}, \bibinfo{person}{Ramchandran Kannan}, {and} \bibinfo{person}{Peter Bartlett}.} \bibinfo{year}{2018}\natexlab{}.
\newblock \showarticletitle{Byzantine-robust distributed learning: Towards optimal statistical rates}. In \bibinfo{booktitle}{\emph{Proceedings of the International Conference on Machine Learning}}. \bibinfo{pages}{5650--5659}.
\newblock


\bibitem[Zhang et~al\mbox{.}(2024)]%
        {c:12}
\bibfield{author}{\bibinfo{person}{Hangfan Zhang}, \bibinfo{person}{Jinyuan Jia}, \bibinfo{person}{Jinghui Chen}, \bibinfo{person}{Lu Lin}, {and} \bibinfo{person}{Dinghao Wu}.} \bibinfo{year}{2024}\natexlab{}.
\newblock \showarticletitle{A3fl: Adversarially adaptive backdoor attacks to federated learning}.
\newblock \bibinfo{journal}{\emph{Advances in Neural Information Processing Systems}}  \bibinfo{volume}{36} (\bibinfo{year}{2024}).
\newblock


\bibitem[Zhang and Zheng(2024)]%
        {c:35}
\bibfield{author}{\bibinfo{person}{Lu Zhang} {and} \bibinfo{person}{Baolin Zheng}.} \bibinfo{year}{2024}\natexlab{}.
\newblock \showarticletitle{FIBA: Federated Invisible Backdoor Attack}. In \bibinfo{booktitle}{\emph{ICASSP 2024-2024 IEEE International Conference on Acoustics, Speech and Signal Processing (ICASSP)}}. IEEE, \bibinfo{pages}{6870--6874}.
\newblock


\bibitem[Zhang et~al\mbox{.}(2021)]%
        {c:17}
\bibfield{author}{\bibinfo{person}{Xinwei Zhang}, \bibinfo{person}{Mingyi Hong}, \bibinfo{person}{Sairaj Dhople}, \bibinfo{person}{Wotao Yin}, {and} \bibinfo{person}{Yang Liu}.} \bibinfo{year}{2021}\natexlab{}.
\newblock \showarticletitle{Fedpd: A federated learning framework with adaptivity to non-iid data}.
\newblock \bibinfo{journal}{\emph{IEEE Transactions on Signal Processing}}  \bibinfo{volume}{69} (\bibinfo{year}{2021}), \bibinfo{pages}{6055--6070}.
\newblock


\bibitem[Zhao et~al\mbox{.}(2022)]%
        {c:7}
\bibfield{author}{\bibinfo{person}{Jianxin Zhao}, \bibinfo{person}{Xinyu Chang}, \bibinfo{person}{Yanhao Feng}, \bibinfo{person}{Chi~Harold Liu}, {and} \bibinfo{person}{Ningbo Liu}.} \bibinfo{year}{2022}\natexlab{}.
\newblock \showarticletitle{Participant selection for federated learning with heterogeneous data in intelligent transport system}.
\newblock \bibinfo{journal}{\emph{IEEE Transactions on Intelligent Transportation Systems}} \bibinfo{volume}{24}, \bibinfo{number}{1} (\bibinfo{year}{2022}), \bibinfo{pages}{1106--1115}.
\newblock


\bibitem[Zhao et~al\mbox{.}(2018)]%
        {c:16}
\bibfield{author}{\bibinfo{person}{Yue Zhao}, \bibinfo{person}{Meng Li}, \bibinfo{person}{Liangzhen Lai}, \bibinfo{person}{Naveen Suda}, \bibinfo{person}{Damon Civin}, {and} \bibinfo{person}{Vikas Chandra}.} \bibinfo{year}{2018}\natexlab{}.
\newblock \showarticletitle{Federated learning with non-iid data}.
\newblock \bibinfo{journal}{\emph{arXiv preprint arXiv:1806.00582}} (\bibinfo{year}{2018}).
\newblock


\bibitem[Zheng et~al\mbox{.}(2022)]%
        {c:26}
\bibfield{author}{\bibinfo{person}{Runkai Zheng}, \bibinfo{person}{Rongjun Tang}, \bibinfo{person}{Jianze Li}, {and} \bibinfo{person}{Li Liu}.} \bibinfo{year}{2022}\natexlab{}.
\newblock \showarticletitle{Data-free backdoor removal based on channel lipschitzness}. In \bibinfo{booktitle}{\emph{Proceedings of the European Conference on Computer Vision}}. \bibinfo{pages}{175--191}.
\newblock


\bibitem[Zhu et~al\mbox{.}(2023b)]%
        {c:25}
\bibfield{author}{\bibinfo{person}{Mingli Zhu}, \bibinfo{person}{Shaokui Wei}, \bibinfo{person}{Li Shen}, \bibinfo{person}{Yanbo Fan}, {and} \bibinfo{person}{Baoyuan Wu}.} \bibinfo{year}{2023}\natexlab{b}.
\newblock \showarticletitle{Enhancing fine-tuning based backdoor defense with sharpness-aware minimization}. In \bibinfo{booktitle}{\emph{Proceedings of the IEEE/CVF International Conference on Computer Vision}}. \bibinfo{pages}{4466--4477}.
\newblock


\bibitem[Zhu et~al\mbox{.}(2023a)]%
        {c:8}
\bibfield{author}{\bibinfo{person}{Rongbo Zhu}, \bibinfo{person}{Mengyao Li}, \bibinfo{person}{Jiangjin Yin}, \bibinfo{person}{Lubing Sun}, {and} \bibinfo{person}{Hao Liu}.} \bibinfo{year}{2023}\natexlab{a}.
\newblock \showarticletitle{Enhanced federated learning for edge data security in intelligent transportation systems}.
\newblock \bibinfo{journal}{\emph{IEEE Transactions on Intelligent Transportation Systems}} \bibinfo{volume}{24}, \bibinfo{number}{11} (\bibinfo{year}{2023}), \bibinfo{pages}{13396--13408}.
\newblock


\end{thebibliography}
\end{document}